\documentclass[twoside]{article}

\usepackage{hyperref}
\usepackage[dvipsnames]{xcolor}
\usepackage{bm}
\usepackage{amsfonts} 
\usepackage{amsmath}
\usepackage{amsthm}
\usepackage{graphicx}

\usepackage[accepted]{aistats2024}
\usepackage{natbib}
\setcitestyle{authoryear} 

\newtheorem{theorem}{Theorem}[section]
\newtheorem{proposition}[theorem]{Proposition}

\newtheorem{definition}[theorem]{Definition}

\newtheorem{remark}[theorem]{Remark}

\AtBeginDocument{
  \let\div\relax
  \DeclareMathOperator{\div}{div}

}

\begin{document}

%
\runningtitle{KuramotoGNN: Reducing Over-smoothing via a Kuramoto Model-based Approach}

%
\runningauthor{Tuan Nguyen, Hirotada Honda, Takashi Sano, Vinh Nguyen, Shugo Nakamura, Tan M. Nguyen}

\twocolumn[

\aistatstitle{From Coupled Oscillators to Graph Neural Networks: Reducing Over-smoothing via a Kuramoto Model-based Approach}

\aistatsauthor{ Tuan Nguyen$^{1,2}$ \And Hirotada Honda$^{2}$ \And  Takashi Sano$^{2}$  }
\aistatsauthor{ Vinh Nguyen$^{*,1}$ \And Shugo Nakamura$^{*,2}$ \And Tan M. Nguyen$^{*,3}$}

\aistatsaddress{} ]

\begin{abstract}
We propose the Kuramoto Graph Neural Network (KuramotoGNN), a novel class of continuous-depth graph neural networks (GNNs) that employs the Kuramoto model to mitigate the over-smoothing phenomenon, in which node features in GNNs become indistinguishable as the number of layers increases. The Kuramoto model captures the synchronization behavior of non-linear coupled oscillators. Under the view of coupled oscillators, we first show the connection between Kuramoto model and Graph Neural Diffusion (GRAND) which is a basic type of GNN and then over-smoothing phenomenon in GNNs can be interpreted as phase synchronization in Kuramoto model. The KuramotoGNN replaces this phase synchronization with frequency synchronization to prevent the node features from converging into each other while allowing the system to reach a stable synchronized state. We experimentally verify the advantages of the KuramotoGNN over the baseline GNNs and existing methods in reducing over-smoothing on various graph deep learning benchmark tasks.
\end{abstract}
\section{Introduction}
\label{sec:intro}

Graph neural networks (GNNs) have been widely adopted in applications such as computational chemistry, social networks, and drug discovery due to their ability to capture complex relationships between nodes and edges in a graph \citep{gilmer2017neural, fan2019graph, xiong2019pushing}. GNNs use multiple graph propagation layers to iteratively update each node's representation by aggregating the representations of its neighbors and the node itself. However, a significant limitation of GNNs is the over-smoothing problem, which occurs when the GNN repeatedly aggregates information from neighboring nodes. This can cause the representations of nodes from different classes to become indistinguishable, leading to reduced model performance \citep{oono, nthoang}.

To address the over-smoothing problem and improve our theoretical understanding of GNNs, recent research has considered GNNs as a discretization scheme of dynamical systems. In this framework, each propagation layer in GNNs is a discrete step of a differential equation \citep{GRAND,GRAND++,GraphCON,CGNN,oono}. This class of models is often referred to as continuous-depth models, and they are more memory-efficient and can effectively capture the dynamics of hidden layers \citep{NODE}.

In this paper, we propose a new physically-inspired framework for understanding GNNs and solving the over-smoothing by using the Kuramoto model \citep{10.1007/BFb0013365}. The Kuramoto model describes the dynamical behavior of nonlinear coupled oscillators and provides a useful proxy for explaining the over-smoothing problem in GNNs. Specifically, we demonstrate that the over-smoothing problem in GNNs is analogous to phase synchronization in coupled oscillators, where all oscillators in the network rotate spontaneously with a common frequency and phase. 
Building on this insight, we propose a new training scheme, called KuramotoGNN, that encourages the model to learn representations that balance between the need to aggregate information from neighboring nodes and the risk of over-smoothing. Our experiments on standard benchmark datasets demonstrate that KuramotoGNN outperforms several baseline models, including popular GNN variants, on node classification tasks. Overall, our work provides a promising direction for addressing the over-smoothing problem in GNNs and opens up new avenues for exploring the connection between GNNs and dynamical systems.

The rest of the paper is organized as follows. Section 2 provides an overview of related work on GNNs. Section 3 provides the background of the continuous-depth GNNs model and Kuramoto model. Section 4 presents our proposed framework KuramotoGNN, together with the similarities to other continuous-depth GNN models and the over-smoothing phenomenon. Section 5 presents our experimental results, and we conclude the paper in Section 6 with a discussion of the implications of our framework and directions for future research.

\section{Related Works}

\textbf{Neural ODEs.} Neural ODEs (NODE) are a class of continuous-depth models for neural networks based on Ordinary Differential Equations. The idea of NODE was first proposed in~\cite{NODE}, and builds on previous studies exploring the relationship between deep learning and differential equations \citep{haber2017stable}. Mathematically, NODE is represented as the following first-order ODE:
\begin{gather}
\label{NODE}
\frac{{\rm d} z(t)}{{\rm d} t} = f(z(t),t, \theta)
\end{gather}
where $f(z(t),t,\theta)$ is specified by a neural network and $\theta$ is its weights. Using numerical methods, such as the Euler discretization with a step size of 1, \eqref{NODE} can be discretized into a vanilla residual network~\citep{residualNet}. The NODE architecture has several advantages in the training process, including the use of the adjoint method \cite{adjoint} for back-propagation, which is more memory-efficient than saving all states of intermediate layers.

Since the proposal of NODE, numerous works have explored the use of ODEs in deep learning for image classification and time-series prediction, such as Neural CDE \citep{NeuralCDE}, Neural SDE \citep{NeuralSDE}, and augmented NODE \citep{augNODE}. Despite the advantages of NODEs, there are also some limitations and challenges associated with their use. For example, it can be difficult to determine the appropriate discretization method for a given problem. Nonetheless, NODEs have demonstrated significant promise as a class of continuous-depth models for neural networks.

\textbf{GNNs.} Graph Neural Networks (GNNs)~\citep{GCN,GAT,GraphSAGE} are a class of models that can effectively learn graph representations. Generally, GNNs have multiple propagation layers, where in each layer, the representation of each node is updated according to representation features of neighboring nodes, called messages. Each type of model has a different type of message aggregation function. However, it has been shown that GNNs are prone to over-smoothing \citep{oono,nthoang}, which occurs when node representations converge to the same values. In particular, it has been shown analytically that deeper GNNs exponentially lose expressive power as the number of layers goes to infinity, and nodes of equal degree in the same connected component will have the same representation \cite{oono}.

Several approaches have been proposed to address this over-smoothing problem. An early solution is concatenation, which is effective, but does not easily scale to very deep networks, due to the size of latent features increasing with the number of layers. Another approach is residual mapping, as used in GCNII~\citep{GCNII}, which maintains good performance up to 64 layers of GNNs. However, in GCNII, a new parameter, called the scale parameter, is introduced to control the level of smoothing, increasing the complexity of the model and making it more difficult to train and optimize.

Recently, inspired by the Neural ODEs~\citep{NODE}, works have turned conventional GNNs into continuous-depth models to solve the over-smoothing problem. For example, \cite{GRAND++,CGNN} added source terms to the linear ODE to change the convergence point of the equation, \cite{GRAND} used heat diffusion-type Partial Differential Equations (PDEs) to design GNNs and, to some extent, slowed the over-smoothing process, and \cite{GraphCON} modeled GNNs as second-order oscillator PDEs with damping terms to analyze the dynamic behavior of the model to avoid over-smoothing. 

The work presented in \cite{GraphCON} (GraphCON) is closest to ours. Both KuramotoGNN and GraphCON share a common objective: addressing the oversmoothing problem inherent in GNNs, while drawing inspiration from coupled oscillators. On the one hand, our work, KuramotoGNN, contributes by establishing a formal connection between oversmoothing and phase synchronization through theoretical analysis. On the other hand, GraphCON's theoretical foundations predominantly revolve around second-order dynamics of coupled oscillators. This framework effectively addresses the oversmoothing by preventing its occurrence under suitable parameter settings. However, GraphCON does not explicitly explore the theoretical linkage between oversmoothing and synchronization, and the model's behavior beyond the oversmoothing context needs further investigation. For instance, whether GraphCON is stable or not in the Lyapunov sense remains a question that requires deeper exploration. Note that although the Kuramoto model is first-order, it can also be extended to second-order ODE equations. In that case, KuramotoGNN and GraphCON are quite similar in the form of equations.

\textbf{Synchronization.} Synchronization is a phenomenon observed in complex networks across many fields, including biology, chemistry, physics, and social systems. One classic example is the synchronous flashing of fireflies, where initially the fireflies flash randomly, but after a short period of time, the entire swarm starts flashing in unison. Coupled oscillator networks are commonly used to study the dynamics of synchronization in networks, where a population of oscillators is connected by a graph that describes the interactions between the oscillators. The oscillator is a simple yet powerful concept that captures a rich dynamic behavior \citep{dorfler2014synchronization}. 


The Kuramoto model has been applied across a wide range of disciplines, including biology, neuroscience, engineering, and even in data processing. For example, the Kuramoto model has been used to study brain networks \citep{varela2001brainweb}, laser arrays \citep{kozyreff2000global}, power grids \citep{motter2013spontaneous}, wireless sensor networks \citep{tanaka2009self}, consensus problems \citep{olfati2007consensus}, and data clustering as a method of unsupervised machine learning \citep{miyano2007data,miyano2008collective}. For further references, we refer to \cite{STROGATZ20001} which provides a comprehensive introduction to synchronization phenomena and their applications, while the review by \cite{dorfler2014synchronization, RevModPhys.77.137} focus on the dynamics of complex networks and synchronization. 

Interestingly, there is a connection between synchronization and the over-smoothing phenomenon in GNNs. Both involve a collective behavior of the nodes in the network, where nodes become more similar to each other over time. By leveraging the insights from the study of synchronization, we can potentially gain a deeper understanding of the over-smoothing problem in GNNs and develop new solutions to address this issue.

\section{Background}
\label{sec:preliminaries}

\subsection{Graph Neural Diffusion (GRAND)}
\label{graph_diffusion}
Graph Neural Diffusion (GRAND) is an architecture for graphs that uses the diffusion process \citep{GRAND}. A graph is defined as $G=(\mathbf{V},\mathbf{E})$, where $\mathbf{V} \in \mathbb{R}^{n \times f}$ represents the $n$ vertices, each with $f$ features, and $\mathbf{E}:=E_{ij}$ is an $n \times n$ matrix that represents the edge weights between the nodes. The model is governed by the following Cauchy problem (we provide detail explanation for equation \eqref{diffusion} in \textbf{SM}).

\begin{gather}
    \label{diffusion}
    \frac{{\rm d} \mathbf{X}(t)}{{\rm d}t} = \div(\mathcal{G}(X(t),t) \odot \nabla \mathbf{X}(t))  \\
    \mathbf{X}(0) = \psi(\mathbf{V})
\end{gather}

where $d$ is the size of encoded input features, $\psi:\mathbb{R}^{n\times f}\rightarrow\mathbb{R}^{n\times d}$ is an affine map that represents the encoder function to the input node features, $\mathbf{X(t)}=[\bigl(\mathbf{x}_1(t)\bigr)^\top, \dots, \bigl(\mathbf{x}_n(t)\bigr)^\top]^\top \in \mathbb{R}^{n \times d}$ is the node function matrix, $\odot$ is the point-wise multiplication, and $\div$ is the divergence operator. 


In the simplest case, when $\mathcal{G}$ is only dependent on the initial value, $\mathbf{X}(0)$, then the equation becomes:
\begin{gather}
    \label{GRAND_eq}
    \frac{{\rm d} \mathbf{X}(t)}{{\rm d}t} = (\hat{\mathbf{A}}-\mathbf{I})\mathbf{X}(t)
\end{gather}
where $\hat{\mathbf{A}}$ is an $n \times n$ matrix, and it is right-stochastic (i.e., each row of $\hat{\mathbf{A}}\odot\mathbf{E}$ summing to 1) which is related to attention weight. In the GRAND model \citep{GRAND}, to formulate the matrix $\hat{\mathbf{A}}$, they used the multi-head self-attention mechanism \citep{attention2017}. The scaled-dot product attention is given by
\begin{gather}
    \label{attention}
    \mathbf{A}^l(\mathbf{X}_i(0), \mathbf{X}_j(0))=\text{softmax}\left(\frac{(\mathbf{W}_K\mathbf{X}_i(0))^\top \mathbf{W}_Q\mathbf{X}_j(0)}{d_k}\right)
\end{gather}
where $d_k$ is a hyper-parameter, and $\mathbf{W}_K$, $\mathbf{W}_Q \in \mathbb{R}^{d_k \times d}$ are the learnable parameters. Then, $\hat{\mathbf{A}}_{ij}=\frac{1}{h}\sum_{l=1}^h\mathbf{A}^l(\mathbf{X}_i(0), \mathbf{X}_j(0))$ with $h$ is the hyper-parameter for the number of heads.


Models derived from \eqref{GRAND_eq} have been shown to extend to a depth larger than conventional GNNs while still maintaining acceptable performance \citep{GRAND}. However, in Section 4, we argue that even with this type of model, over-smoothing cannot be completely eliminated.

\subsection{The Kuramoto model}
In the study of interacting limit-cycle oscillators, weakly coupled systems have been found to exhibit dynamical behavior that depends only on the phases of the oscillators, as demonstrated by \cite{WINFREE196715}. Such models are called phase-reduced models. One of the most well-known phase-reduced models is the Kuramoto model, which describes the dynamics of a network of $N$ phase oscillators $\theta_i$, with natural frequencies $\omega_i$ and coupling strength $\kappa_{ij}$, using the following phase equation:

\begin{gather}
\dot{\theta_i} = \omega_i + \sum_{j}\kappa_{ij}\sin(\theta_j-\theta_i)
\label{general_kuramota}
\end{gather}

Each oscillator, $i$, is characterized by intrinsic and extrinsic factors \citep{RevModPhys.77.137}. The internal influence is the natural frequency of the oscillator, $\omega_i$, while the external influences are the interactions with other oscillators in the network through the weight (or coupling) matrix, $\kappa \in \mathbb{R}^{n \times n}$. From \eqref{general_kuramota}, the synchronization behavior has been further analyzed using the mean field coupling, that is, $\kappa_{ij}=K/N$, where $K$ is a constant and $N$ is the number of nodes in the graph \citep{10.1007/BFb0013365}. Hence, \eqref{general_kuramota} becomes:
\begin{align}
\dot{\theta_i} = \omega_i + \frac{K}{N}\sum_{j}\sin(\theta_j-\theta_i)
\label{usual_kuramoto}
\end{align}

Equation \eqref{usual_kuramoto} is the well-known \textit{Kuramoto model}. The Kuramoto model has two types of states: a nonsynchronized state, in which each oscillator rotates independently with its own frequency, and a partially synchronized state, in which some of the oscillators rotate with the same effective frequency. It has been found that strengthening the couplings provides a synchronization transition from the nonsynchronized state to the partially synchronized state and that the continuity of the transition is determined by the natural frequency distribution \citep{10.1007/BFb0013365}. Specifically, if the distribution of natural frequencies $\omega_i$ is unimodal and symmetric, then synchronization occurs in equation \eqref{usual_kuramoto} if the coupling parameter $K$ exceeds a certain threshold $K_{critical}$ determined by the distribution of $\omega_i$.


To investigate synchronization behavior or measure synchronization rate, the author in \cite{10.1007/BFb0013365} introduced a complex-valued term called the \textit{order parameter}:

\begin{equation}
    re^{i\phi} = \frac{1}{N}\sum_{j}e^{i\theta_j}
    \label{order_parameter}
\end{equation}

where $\phi$ is defined as the average phase. $r(t)$ is a real value in the range of $0\leq r(t)\leq 1$, where $r(t) \approx 1$ indicates that the phases are close together or that the synchronization rate is high, while a value of $r(t) \approx 0$ indicates that the phases are spread out over the circle.

\section{KuramotoGNN: Kuramoto Graph Neural Network}

\subsection{Model Formulation}
In this work, we consider the generalized form of the Kuramoto model:
\begin{eqnarray}\label{gather}
    \dot{\theta_i} = \omega_i + \frac{K}{\sum_{j}a_{ij}}\sum_{j}a_{ij}\sin(\theta_j-\theta_i) \label{kuramoto}
\end{eqnarray}

By adapting the setting of initial value and the right-stochastic matrix $\hat{\mathbf{A}}=[a_{ij}]$ which has been stated in \ref{graph_diffusion}, we now can present the KuramotoGNN, a class of continuous-depth GNNs based on the Kuramoto model \eqref{kuramoto}:
\begin{gather}
    \label{KuramotoGNN}
    \dot{\mathbf{x}_i} = \omega_i + K\sum_{j}a_{ij}\sin(\mathbf{x}_j-\mathbf{x}_i)\\
    \mathbf{X}(0)=\Omega=\psi(\mathbf{V})
\end{gather}

with natural frequencies $\omega_i \in \mathbb{R}^d$, $\Omega=[\omega_1^\top,\dots,\omega_n^\top]^\top$, $i^{th}$ node vector representations $\mathbf{x}_i(t) \in \mathbb{R}^d$, and $\sin$ function operates in a component-wise manner. Unlike many previous works related to the Kuramoto model, in which they try to find the threshold $K_{critical}$ to control the dynamics of the system, we approach the problem in a data-driven way. We fix $K$ as a hyperparameter and optimize $\Omega$, $\mathbf{X}(0)$, and $\hat{\mathbf{A}}$ through the training process. We adopt the setting of GRAND, where $a_{ij}$ serves as learnable parameters but depends solely on the initial value $\mathbf{X(0)}$, as shown in equation \eqref{attention}.

\begin{proposition}
    \label{GRAND_is_Kuramoto}
    Graph Neural Diffusion \eqref{GRAND_eq} is the linearized dynamics of the Kuramoto model.
\end{proposition}
\begin{proof}
    Assuming that $\omega_i$ are identical, in which $\omega_1=...=\omega_N$, we obtain the following equation by using the rotating frame of reference:
    \begin{gather}
        \label{omega_same}
        \dot{\mathbf{x}_i} = K\sum_{j}a_{ij}\sin(\mathbf{x}_j-\mathbf{x}_i)
    \end{gather}

    Next, we assume that $K=1$ and roughly approximate the nonlinear $\sin$ function by using the first order of the Taylor expansion, then we can obtain the following equation which is identical to Graph Neural Diffusion (GRAND) \eqref{GRAND_eq}:
    \begin{gather}
        \label{kuramoto_grand}
        \dot{\mathbf{x}_i} = \sum_{j}a_{ij}(\mathbf{x}_j-\mathbf{x}_i)
    \end{gather}
    
    Hence, GRAND \eqref{GRAND_eq} is the linearized dynamics of the Kuramoto model.     
\end{proof}
\begin{remark}
    In GRAND, if $\hat{\mathbf{A}}$ in equation \eqref{GRAND_eq} depends only on the initial value $\left(\mathbf{x}_i(0), \mathbf{x}_j(0)\right)$, then we have a linear version of GRAND named \textbf{GRAND-l}. Meanwhile, if $\hat{\mathbf{A}}$ is time dependent and depends on $\left(\mathbf{x}_i(t), \mathbf{x}_j(t)\right)$, we obtain the non-linear GRAND called \textbf{ GRAND-nl}. On the contrary, the KuramotoGNN model, in general, is nonlinear even in the case that we consider $a_{ij}$ to depend solely on initial value or to depend on time-dependent values.      
\end{remark}

\textbf{On the exploding and vanishing gradients.} In the case of highly deep GNN architectures, it is crucial to explore potential approaches to alleviate the issues of exploding and vanishing gradients. For simplicity and without any loss of generality, we consider the case of the fully-connected graph and scalar node features by setting $d=1$, or $x_i^t=\mathbf{x}_{i,1}(t)$, and explicitly reduce the KuramotoGNN to the Euler discretization form with a step-size $\Delta t \ll 1$,
\begin{gather}
    x_i^t = x_i^{t-1}+\Delta t \left( x_i^0 +\frac{1}{n}\sum \sin(x_j^{t-1}-x_i^{t-1}) \right)\\
    \mathbf{X}(0) = [x_1^0,\dots,x_n^0]^\top = \psi(\mathbf{V}) = \mathbf{V}\mathbf{W}
\end{gather}
Where $W$ is the learnable parameters, $M$ is the number of ODE integrations or number of model's layers, $t=1,\dots,M$. Furthermore, let us consider a scenario where the objective of the GNN is to approximate the ground truth vector $\hat{\mathbf{X}}\in \mathbb{R}^n$ with the following loss function.
\begin{equation}
    \label{loss_func}
    J(W)=\frac{1}{2n}\sum_{i=1}^N\|x_i^n-\hat{x}_i\|^2
\end{equation}
At every step of gradient descent, we need to compute the gradient $\frac{\partial J}{\partial \mathbf{W}}$ which measures the contribution made by parameters $\mathbf{W}$.  Using the chain rule, we obtain the following.
\begin{gather}
    \label{total_grad}
    \frac{\partial J}{\partial \mathbf{W}} = \frac{\partial J}{\partial \mathbf{Z}^L}\frac{\partial {\mathbf{Z}}^L}{\partial \mathbf{Z}^1}\frac{\partial {\mathbf{Z}}^1}{\partial \mathbf{Z}^0}\frac{\partial {\mathbf{Z}}^0}{\partial \mathbf{W}}\\
    \label{long_prod}
    \frac{\partial {\mathbf{Z}}^L}{\partial \mathbf{Z}^1}=\prod_{i=1}^L \frac{\partial {\mathbf{Z}}^i}{\partial \mathbf{Z}^{i-1}}
\end{gather}
Here, $\mathbf{Z}^L=[x_1^L,\dots,x_n^L]$ and $L$ is represented as the number of layers. Assuming an approximate relationship $\frac{\partial {\mathbf{Z}}^i}{\partial \mathbf{Z}^{i-1}} \approx \lambda$, the repeated multiplication in equation \eqref{long_prod} implies that $\frac{\partial {\mathbf{Z}}^L}{\partial \mathbf{Z}^1} \approx \lambda^L$. When $\lambda > 1$, the total gradient \eqref{total_grad} can \textit{grow exponentially} with the number of layers, leading to the problem of exploding gradients. In contrast, when $\lambda < 1$, the total gradient \eqref{total_grad} can \textit{decay exponentially} with the number of layers, resulting in the problem of vanishing gradients. These scenarios can hinder successful training as the gradient either becomes excessively large or remains stagnant, impeding effective parameter updates.

\begin{proposition}
    We assume that $\Delta t \ll 1$ is chosen to be sufficiently small. Then, the gradient of the loss function \eqref{loss_func} with respect to any learnable weight parameter $\mathbf{W}$ is bounded as:
    \begin{gather}
        \label{upper_bound}
        \|\frac{\partial J}{\partial \mathbf{W}}\|_\infty \leq \frac{1}{n}\left[\alpha(\max|x_i^0|+1)+\max|\hat{x}_i|\right](\beta+\alpha)\beta\|\mathbf{V}\|_\infty\\
    \alpha = M\Delta t, \quad \quad \beta = 1+\frac{\Delta t}{n}
    \end{gather}
\end{proposition}
Where $\|\mathbf{x}\|_\infty:=\max_i|\mathbf{x}_i|$ is the infinity norm.The upper bound presented in equation \eqref{upper_bound} demonstrates that the total gradient remains globally bounded, regardless of the number of layers $M$, effectively addressing the issue of exploding gradients. However, it is important to note that this upper bound does not automatically eliminate the possibility of vanishing gradients. To further investigate this matter, we follow \cite{GraphCON} to derive the following proposition for the gradients (proof provided in the \textbf{SM}).

\begin{proposition}
     We assume that $\Delta t \ll 1$ is chosen to be sufficiently small. Then, the gradient of the loss function \eqref{loss_func} can be represented as:
     \begin{equation}
        \label{vanishing_eq}
         \frac{\partial J}{\partial \mathbf{W}}=\frac{\partial J}{\partial \mathbf{Z}^M}\left[\Delta t(\mathbf{E'}+\sum_i^M\mathbf{E}_{i-1})+\mathbf{I}+\mathcal{O}(\Delta t^2)\right]\mathbf{W}
     \end{equation}
     with $\frac{\partial J}{\partial \mathbf{Z}^M} = \frac{1}{n}[x_1^M -\hat{x}_1,\dots,x_n^M -\hat{x}_n]$ and the order notation, $\mathbf{E}$, $\mathbf{E'}$, and $\frac{\partial J}{\partial \mathbf{Z}^M}$ are defined in \textbf{SM}.
\end{proposition}
Thus, although the gradient \eqref{vanishing_eq} can be small, \textit{it will not vanish exponentially} by increasing the number of layers $M$, mitigating the vanishing gradient problem.

\textbf{On the stability}. In the Kuramoto model, the stability of the synchronized state if the relative frequency differences converge as $t\rightarrow \infty$, and hence if $\dot{\mathbf{x}}_i$ in \eqref{KuramotoGNN} satisfy the following condition, we can say that the model is stable.
\begin{equation}
    \label{stable_condition}
    \lim_{t\rightarrow \infty}\|\dot{\mathbf{x}}_i-\dot{\mathbf{x}}_j\|=\mathbf{0}, \quad \forall i,j \in \mathcal{V}
\end{equation}
In numerical simulations, it has been observed that the frequency synchronization of Kuramoto oscillators tend to zero, regardless of initial configuration of $\mathbf{x}_i(0)$. One notable advantage of utilizing the Kuramoto model in our study is the extensive availability of previous research papers, which contribute to a rich repository of results, theorems, and analyses. The following theorem help us to understand the stability of \eqref{KuramotoGNN} concretely.

\begin{theorem}
    \label{theo_stable}
    \citep{ha2016emergence} Suppose that the initial configuration $\mathbf{X}(0)$ and natural frequencies $\Omega$ satisfy
    \begin{eqnarray}
        r_0>0, \quad \mathbf{x}_i(0)\neq\mathbf{x}_j(0) \quad \forall i,j \in \mathcal{V}, \quad \max\|\Omega\|<\infty
    \end{eqnarray}
    Then there exists a large coupling strength $K_\infty>0$ such that if $K>K_\infty$ then there exists the stable state \eqref{stable_condition} which is the solution of \eqref{KuramotoGNN} with initial data $\mathbf{X}(0)$.
\end{theorem}
\begin{remark}
    In this stability section, we also consider the case of the scalar node feature for simplicity.
\end{remark}
\begin{remark}
    $r_0$ is the order parameter \eqref{order_parameter} of the initial configuration $\mathbf{X}_0$,  $r_0=|r|=\frac{1}{N}\sum_je^{i\mathbf{x}_j(0)}$. Theorem \ref{theo_stable} covers all initial
    configurations, except that with $r_0=0$ or when all the initial oscillators are distributed uniformly around the circle. 
\end{remark}

\subsection{Over-smoothing as synchronization}
\begin{definition}
    \label{def_over-smoothing}
    The over-smoothing phenomenon occurs when the following condition converges \textit{exponentially} to zero:
    \begin{eqnarray}\label{syns}
        \lim_{t\to T}\|\mathbf{x}_i(t)-\mathbf{x}_j(t)\|=0, \forall i \neq j.
    \end{eqnarray}
\end{definition}

With $T$ is the terminal time of the ODE. In conventional GNN architectures, over-smoothing has been a widely observed and analyzed phenomenon \citep{oono,nthoang}. The previous reference showed that over-smoothing is a phenomenon in which the node representations converge to a fixed state exponentially and, in that fixed state, the node representations are indistinguishable. We notice a strong resemblance between over-smoothing and \textit{phase synchronization} in coupled oscillator dynamics.

In general, oscillator synchronization occurs when they reach the \textit{frequency synchronization} state. In other words, when the frequencies of the coupled oscillators converge to some common frequency, $\|\dot{\mathbf{x}_i}-\dot{\mathbf{x}_j}\|=0,\forall i,j=1,\dots,N$, despite differences in the natural frequencies of the individual oscillators. If, in addition to frequency synchronization, the oscillator representations $\mathbf{x}_i(t)$ converge to a common value, $\|\mathbf{x}_i(t)-\mathbf{x}_j(t)\|=0$, it is called \textit{phase synchronization}.

\begin{remark}
    The definition of the \textit{frequency synchronization} state aligns precisely with equation \eqref{stable_condition}.
\end{remark}

\begin{proposition}
    \label{exp_sync}
    Over-smoothing is the phase synchronization state of the node features. Under the analysis of synchronization, conventional Graph Neural Diffusion is easily drawn into the over-smoothing phenomenon.
\end{proposition}

\begin{proof}
    We first show that the phase synchronization state is an exponentially stable solution of identical Kuramoto oscillators \eqref{omega_same}, then with the Definition \ref{def_over-smoothing} we can indicate that the over-smoothing state is exactly the same as the phase synchronization state. For simplicity, here we consider the scalar node feature, $d=1$, and denote $x_i=\mathbf{x}_{i,1}$. 

    Let us say that if a solution of \eqref{omega_same} achieves phase synchronization, then it does so with a value equal to $x_{sync}(t)$, that is, $x_1(t)=x_2(t)=\dots=x_N(t)=x_{sync}(t)$. By transformation into a rotating frame with frequency $x_{sync}$, we can obtain $x_1(t)=x_2(t)=\dots=x_N(t)=0$ or $\mathbf{X}=0$ if we consider $\mathbf{X}=[x_1^\top(t),\dots,x_N^\top(t)]$.
    
    
    We consider the following energy function $U$:
    \begin{gather}
        \label{energy_func}
        U(\mathbf{X}) = \sum_{{i,j} \in E}a_{ij}(1-\cos(x_i - x_j))
    \end{gather}
    Because $\frac{\partial U}{\partial x_i} = \sum_{j}a_{ij}\sin(x_i - x_j)$, we can represent $\nabla{U(\mathbf{X})} = [\frac{\partial U}{\partial x_1},...,\frac{\partial U}{\partial x_n}]$ as follows:
    \begin{gather}
        \label{deriv_energy}
        \nabla{U(\mathbf{X})} = -\frac{\dot{\mathbf{X}}^{\top}}{K}
    \end{gather}
    Which leads to the time derivative of $U$:
    \begin{equation}
        \dot U = \nabla{U(\mathbf{X})} \dot{\mathbf{X}} = -\frac{1}{K}\dot{\mathbf{X}}^\top\dot{\mathbf{X}} \leq 0
    \end{equation}
    
    Using the LaSalle Invariance Principle \citep[Theorem 4.4]{LaSelle}, which is akin to the Lyapunov method, we can analyze the behavior of all solutions of an autonomous ODE as $t\rightarrow \infty$. The principle gives conditions that describe the behavior of the system rather than focusing on the stability of a particular equilibrium solution as in the Lyapunov method. Specifically, if a positive and non-increasing scalar-valued function $U: \mathbf{R}^n \rightarrow \mathbf{R}$ (also known as a Lyapunov or energy function) satisfies $\dot{U}(x) \leq 0$ for all $x$, then every solution converges exponentially to a set of critical points $\{x | \dot{U}(x) = 0\}$.
    
    Therefore, following LaSalle Invariance Principle, every solution of \eqref{omega_same} converges exponentially to a set of critical points that are the root of the right-hand side of \eqref{omega_same}. Looking at \eqref{omega_same}, it is easily confirmed that the phase synchronization state, $x_i=x_j, \forall i \neq j$, is a solution of the equation. Furthermore, it has been shown that the phase synchronization state is the only global attractor, and the synchronization time scales with the inverse of the smallest non-zero eigenvalue of the Laplacian matrix \citep{arenas2008synchronization}. Therefore, \textit{the phase synchronization state is the only exponentially stable state} of equation \eqref{omega_same}. And now, together with the Definition \ref{def_over-smoothing}, we can clearly see that over-smoothing is the phase synchronization state of the node features.
    
    We have demonstrated the equivalence between over-smoothing and phase synchronization. We have also shown that in the scenario of a single channel of node representations and identical oscillators (as in equation \eqref{omega_same}), phase synchronization occurs exponentially. For multichannel or when $d>1$, the same conclusion can be derived in a similar way for each channel. And thus, along with Proposition \ref{GRAND_is_Kuramoto} we can say that under the analysis of synchronization, conventional Graph Neural Diffusion \eqref{kuramoto_grand} is likely encountered with the over-smoothing phenomena.
\end{proof}


We next present a theorem that offers an easy way to reduce the over-smoothing phenomenon by introducing non-identical natural frequencies $\omega_i$ into the equation.  
\begin{theorem}\label{main}
Let $\mathbf{x}_i(.), i=1,\dots,N$ be a solution of~~  \eqref{KuramotoGNN}. If there exists $i,j \in 1,\dots,N$ such that $\omega_i \neq \omega_j$, then \eqref{syns} does not happen.
\end{theorem}





We explain how our model can avoid over-smoothing with the help of Theorem~~\ref{main}. Recall that we set $\omega_i = \psi(\mathbf{V}_i)=\mathbf{D}\mathbf{V}_i+\mathbf{b} \in \mathbb{R}^d$, where $\psi$ is a learnable function among linear functions, and $\mathbf{\mathbf{V}_i}$ is the input feature vector of the $i^{th}$ node, and there are a total of $N$ distinct nodes. It is obvious that $\mathbf{V}_i \neq \mathbf{V}_j, \forall i\neq j$. And thus, by passing through  $\psi$,  $\omega_i=\omega_j, \forall i\neq j$ or $\psi(\mathbf{V}_i)=\psi(\mathbf{V}_j), \forall i\neq j$ is equivalent to saying $D(\mathbf{V}_i - \mathbf{V}_j)=0, \forall i\neq j$, which is impossible because the number of row rank of $\mathbf{D} \leq N\ll \frac{N(N-1)}{2}$, with $\frac{N(N-1)}{2}$ is the number of pair $\mathbf{V}_i\neq \mathbf{V}_j$.

\begin{remark}
    Nonidentical frequencies $\omega_i$ in equation \eqref{KuramotoGNN} prevent phase synchronization, but according to Theorem \ref{theo_stable}, a stable frequency synchronization state can still be achieved. Thus, nonidentical frequencies induce a transition from phase synchronization to frequency synchronization.
\end{remark}

\begin{remark}
    \cite{oono} highlighted the occurrence of over-smoothing in conventional GNN architectures, where node representations converge to the over-smoothing state exponentially. Interestingly, this phenomenon is strongly similar to the \textit{synchronization manifold} in coupled oscillator dynamics. In synchronization, the oscillators evolve synchronously on the same solution, $\lim_{t\to\infty}\mathbf{x}_1(t)=\dots=\mathbf{x}_N(t)$, which is defined by the eigenvector with the lowest eigenvalue, $\lambda_1$. Similarly, through their work, over-smoothing happens when the dynamics of node representations approach an invariant subspace that corresponds to the lowest frequency of graph spectra. Therefore, it is apparent that the concept of synchronization can be used to understand the phenomenon of over-smoothing in GNNs.
\end{remark}

\section{EXPERIMENTAL RESULTS}

We conduct experiments to compare the performance of our proposed method, KuramotoGNN, with GRAND, GRAND++, GCNII, GraphCON and other popular GNN architectures on node classification tasks, including GCN, GAT, and GraphSage. For all experiments, we run 100 splits for each dataset with 20 random seeds for each split, and we conducted on a server with one NVIDIA RTX 3090 graphics card. 

For most of the settings, we adopt from GRAND \citep{GRAND} for KuramotoGNN including adaptive numerical differential equation solvers. Except for 2 factors: the coupling hyper-parameter, which belongs solely to the KuramotoGNN, and the integration time, which measures the implicit depth of continuous-model, were slightly changed. Following~\cite{GRAND}, we study seven graph node classification datasets, namely CORA \citep{Cora}, CiteSeer \citep{Citeseer}, PubMed \citep{Pubmed}, CoauthorCS \citep{Coauthor}, the Amazon co-purchasing graphs Computer and Photo \citep{ComPhoto}. The descriptions of experimental settings in the \textbf{SM}. 


\begin{figure*}[!ht]
  \centering
    \includegraphics[width=\textwidth]{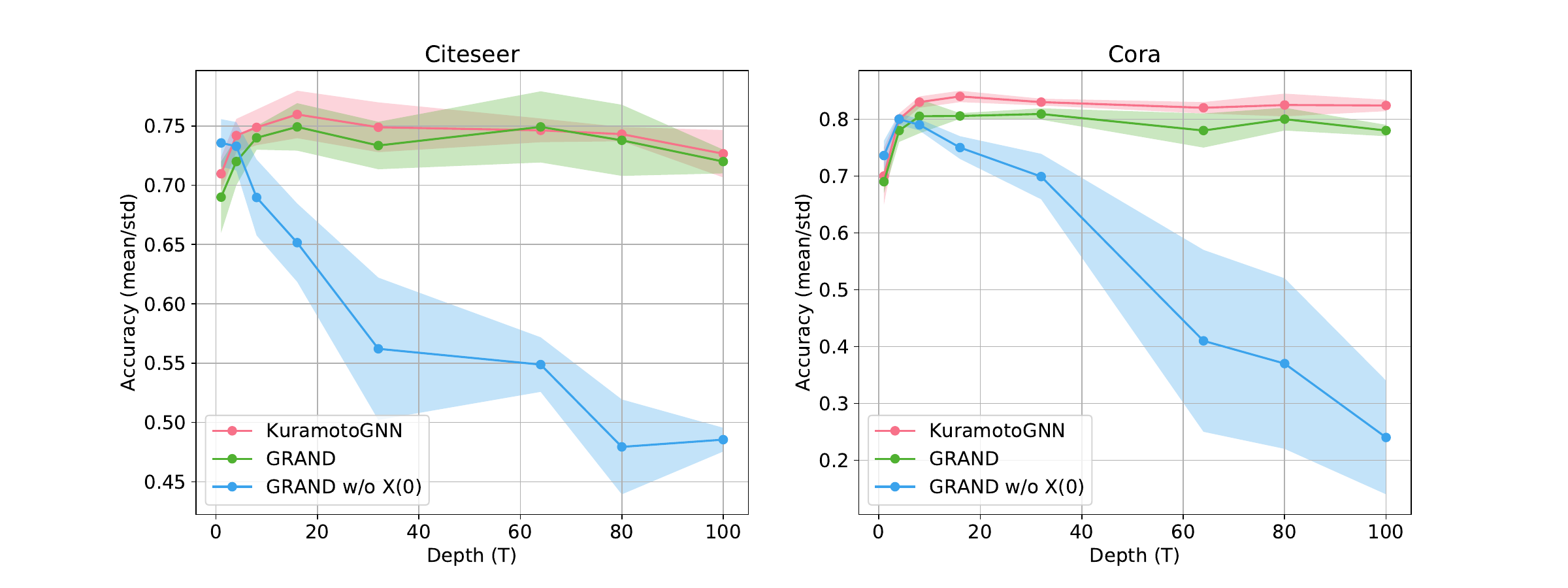}
    \centering
    \caption{Change in performance at different depth ($T$) on Cora and Citeseer dataset.}
  \label{fig:depth}    
\end{figure*}


\begin{table*}[t]
\centering
    \caption{Mean and std of classification accuracy of KuramotoGNN and other GNNs on six benchmark graph node classification tasks. The highest accuracy is highlighted in bold for each number of labeled data per class. (Unit: \%)}
    \resizebox{\textwidth}{!}{\begin{tabular}{cccccccc}
         \hline
          Model & CORA & Citeseer & Pubmed & Computers & CoauthorCS & Photo \\  
         \hline 
         KuramotoGNN & \textbf{85.18$\pm$1.3} & \textbf{76.01$\pm$1.4} & \textbf{80.15$\pm$0.3} & \textbf{84.6$\pm$0.59} & \textbf{92.35$\pm$0.2} & \textbf{93.99$\pm$0.17} \\
         \hline 
         GraphCON& 84.2$\pm$1.3 & 74.2$\pm$1.7 & 79.4$\pm$1.3 & 84.1$\pm$0.9 & 90.5$\pm$1.0& 93.16$\pm$0.5 \\
         \hline 
         GRAND++& 82.95$\pm$1.37 & 73.53$\pm$3.31 & 79.16$\pm$1.37 & 82.99$\pm$0.81 & 90.80$\pm$0.34& 93.55$\pm$0.38 \\
        \hline
         GRAND-l& 82.46$\pm$1.64 & 73.4$\pm$5.05 & 78.8$\pm$1.63 & 84.27$\pm$0.6 & 91.24$\pm$0.4& 93.6$\pm$0.4 \\
        \hline
         GCNII& 84.02$\pm$0.5 & 70.26$\pm$0.7 & 78.95$\pm$0.9 & 80.28$\pm$2.1 & 91.11$\pm$0.2& 92.1$\pm$0.4 \\
        \hline
        GCN& 82.07$\pm$2.03 & 74.21$\pm$2.90 & 76.89$\pm$3.27 & 82.94$\pm$1.54 & 91.09$\pm$0.35 & 91.95$\pm$0.11 \\
        \hline
        GAT& 80.04$\pm$2.54 & 72.02$\pm$2.82 & 74.55$\pm$3.09 & 79.98$\pm$0.96 & 91.33$\pm$0.36 & 91.29$\pm$0.67 \\
        \hline
        GraphSAGE & 82.07$\pm$2.03 & 71.52$\pm$4.11 & 76.49$\pm$1.75 & 73.66$\pm$2.87 & 90.31$\pm$0.41 & 88.61$\pm$1.18 \\
        \hline
    \end{tabular}}
    \label{tab:result_acc}
\end{table*}

\subsection{KuramotoGNN is resilient to deep layers}
To demonstrate that the model does not suffer from over-smoothing, we conducted experiments in various of depth (by changing the integration limit, $T$) and measure the performance in accuracy. One notable point is that the proposed equation in the GRAND paper \eqref{GRAND_eq} was modified in its implementation, as described below:

\begin{gather}
    \label{modified_GRAND}
    \frac{{\rm d} \mathbf{X}(t)}{{\rm d}t} = \alpha(\hat{\mathbf{A}}-\mathbf{I})\mathbf{X}(t)+\beta \mathbf{X}(0)
\end{gather}
with learnable $\alpha$ and $\beta$ parameters. 
In fact, \eqref{modified_GRAND} bears a striking resemblance to our proposed \eqref{KuramotoGNN}, if we make a rough approximation of the function $\sin$ as previously mentioned.

In this experiment, we conduct both versions of linear GRAND, with and without adding $\mathbf{X}(0)$. For all models, we used the random split method with 10 initialization, along with the Euler step-fixed solver with step size 0.1 step size for a fair comparison in the computational process instead of using an adaptive step size scheme which gives more superior results~\citep{GRAND}. 

Figure \ref{fig:depth} shows the change in accuracy of three kinds of models: \textbf{KuramotoGNN}, \textbf{GRAND-l} and \textbf{GRAND-l w/o X(0)} for various depth values $T=\{1, 4, 8, 16, 32, 64, 80, 100\}$. We can see that without adding $\mathbf{X}(0)$, the performance of \textbf{GRAND-l} is reduced significantly along the depth, while the \textbf{KuramotoGNN} and \textbf{GRAND-l w/o X(0)} maintain the performance when increasing depth. 


\subsection{KuramotoGNN performances on various benchmarks}
We evaluate the effectiveness of our model on six popular graph benchmarks. Our KuramotoGNN demonstrates better performance in
terms of accuracy when compared to other continuous models: GraphCON\citep{GraphCON}, GRAND++\citep{GRAND++}, GRAND-l\citep{GRAND}, GCNII\citep{GCNII}, and traditional models: GCN\citep{GCN}), GAT\citep{GAT}, and GraphSAGE\citep{GraphSAGE}.. Table \ref{tab:result_acc} compares the accuracy of fine-tuned Kuramoto with GRAND-l (following equation \eqref{modified_GRAND}) and other GNNs.  


\section{Conclusion}
In summary, we introduce the Kuramoto Graph Neural Network (KuramotoGNN), a novel class of continuous-depth graph neural networks inspired by the Kuramoto model. Our theoretical analysis establishes a connection between the over-smoothing problem and \textit{phase synchronization} in coupled oscillator networks. By incorporating non-identical natural frequency terms, we mitigate the over-smoothing issue, leveraging insights from synchronization and the Kuramoto model. Empirical experiments demonstrate the superior performance of KuramotoGNN compared to other GNN variants, especially with limited labeled data. It is worth noting that our work focuses on the classic Kuramoto model, while future research could explore variations such as time-delayed Kuramoto, adaptive coupling functions, or second-order Kuramoto with damping \citep{dorfler2014synchronization}.

\subsubsection*{Acknowledgements}
This work was supported by Toyo University Top Priority Research Program.

\bibliography{references}

 \begin{enumerate}

\section*{Checklist}

 \item For all models and algorithms presented, check if you include:
 \begin{enumerate}
   \item A clear description of the mathematical setting, assumptions, algorithm, and/or model. [\textcolor{green}{Yes}/No/Not Applicable]
   \item An analysis of the properties and complexity (time, space, sample size) of any algorithm. [Yes/\textcolor{green}{No}/Not Applicable]
   \item (Optional) Anonymized source code, with specification of all dependencies, including external libraries. [Yes/No/Not Applicable]
 \end{enumerate}

 \item For any theoretical claim, check if you include:
 \begin{enumerate}
   \item Statements of the full set of assumptions of all theoretical results. [\textcolor{green}{Yes}/No/Not Applicable]
   \item Complete proofs of all theoretical results. [\textcolor{green}{Yes}/No/Not Applicable]
   \item Clear explanations of any assumptions. [\textcolor{green}{Yes}/No/Not Applicable]     
 \end{enumerate}

 \item For all figures and tables that present empirical results, check if you include:
 \begin{enumerate}
   \item The code, data, and instructions needed to reproduce the main experimental results (either in the supplemental material or as a URL). [\textcolor{green}{Yes}/No/Not Applicable]
   \item All the training details (e.g., data splits, hyperparameters, how they were chosen). [\textcolor{green}{Yes}/No/Not Applicable]
         \item A clear definition of the specific measure or statistics and error bars (e.g., with respect to the random seed after running experiments multiple times). [\textcolor{green}{Yes}/No/Not Applicable]
         \item A description of the computing infrastructure used. (e.g., type of GPUs, internal cluster, or cloud provider). [\textcolor{green}{Yes}/No/Not Applicable]
 \end{enumerate}

 \item If you are using existing assets (e.g., code, data, models) or curating/releasing new assets, check if you include:
 \begin{enumerate}
   \item Citations of the creator If your work uses existing assets. [\textcolor{green}{Yes}/No/Not Applicable]
   \item The license information of the assets, if applicable. [Yes/No/\textcolor{green}{Not Applicable}]
   \item New assets either in the supplemental material or as a URL, if applicable. [Yes/No/\textcolor{green}{Not Applicable}]
   \item Information about consent from data providers/curators. [Yes/No/\textcolor{green}{Not Applicable}]
   \item Discussion of sensible content if applicable, e.g., personally identifiable information or offensive content. [Yes/No/\textcolor{green}{Not Applicable}]
 \end{enumerate}

 \item If you used crowdsourcing or conducted research with human subjects, check if you include:
 \begin{enumerate}
   \item The full text of instructions given to participants and screenshots. [Yes/No/\textcolor{green}{Not Applicable}]
   \item Descriptions of potential participant risks, with links to Institutional Review Board (IRB) approvals if applicable. [Yes/No/\textcolor{green}{Not Applicable}]
   \item The estimated hourly wage paid to participants and the total amount spent on participant compensation. [Yes/No/\textcolor{green}{Not Applicable}]
 \end{enumerate}

 \end{enumerate}

\end{document}


%
\runningtitle{KuramotoGNN: Reducing Over-smoothing via a Kuramoto Model-based Approach}

%
\runningauthor{Tuan Nguyen, Hirotada Honda, Takashi Sano, Vinh Nguyen, Shugo Nakamura, Tan M. Nguyen}

\onecolumn
\aistatstitle{Supplement to ``From Coupled Oscillators to Graph Neural Networks: Reducing Over-smoothing via a Kuramoto Model-based Approach''}

\vspace{-2in}
\tableofcontents

\newpage
\section{FURTHER DISCUSSION}
\subsection{THE NEED OF ALLEVIATING OVERSMOOTHING}

We believe that addressing oversmoothing can bring significant benefits to constructing very deep models and handling limited labeled training data. When faced with the challenge of limited labeled training data, the key issue is extracting meaningful features that effectively capture the underlying graph patterns and nuances. Oversmoothing exacerbates this challenge by leading to information loss, which, in turn, reduces the model's ability to accurately differentiate between nodes \citep{calder2020poisson}. Our intuition suggests that mitigating oversmoothing can enable the model to capture and retain relevant features that might otherwise remain obscured. This enhancement strengthens the model's capacity to learn distinctive representations, even when working with a limited amount of labeled data under the realizability assumption, where the true function falls within the hypothesis set we consider.

The ability to construct deeper graph neural networks provides versatile opportunities across various applications. While deeper KuramotoGNNs do not inherently guarantee better performance for specific tasks, the flexibility to create models with varying depths is a significant advantage. This flexibility is particularly valuable because it opens the door to broader applicability across other neural ODE techniques. Furthermore, our observation indicates that graph neural networks have been successfully employed for learning complex dynamical systems \citep{pfaff2020learning}. Consequently, the ability to build continuous-depth graph neural networks suitable for large-time $T$ holds substantial promise for studying the long-term behavior of complex physical systems.

\subsection{FURTHER COMPARISONS TO RELATED MODELS}

In this context, we provide additional comparisons involving GRAND (a representative linear ODE model) and GraphCON (a model closely related to ours).

While the Kuramoto model is inherently a first-order ODE, it can also be extended to second-order ODE equations. In this case, the equations for KuramotoGNN and GraphCON exhibit a notable similarity in their form.

For the 2nd order KuramotoGNN, the equations are as follows:

\begin{align*}
    \dot{x}_i &= y_i\\
    \dot{y}_i& =\sum_{j \in N(i)}a_{ij}\sin(x_j-x_i) - \omega_i - \alpha y_i
\end{align*}

On the other hand, GraphCON's equations are expressed as:

\begin{align*}
    \dot{x}_i &= y_i\\
    \dot{y}_i& =\sigma(\sum_{j \in N(i)} a_{ij} x_j) - x_i - \alpha y_i
\end{align*}
One can notice that the differences are the coupling function and $\omega,X_i$ terms. In GraphCON, they consider the $\sigma$ function as the ReLU function.

Another notable distinction between GRAND and the Kuramoto model pertains to their mathematical nature. GRAND is a linear ODE model, whereas the Kuramoto model is inherently nonlinear. This nonlinearity imparts KuramotoGNN with a richer and more expressive dynamic behavior compared to GRAND. For instance, while GRAND may have just one equilibrium point, the Kuramoto model can exhibit multiple stable solutions that extend beyond the confines of an equilibrium point. Instead, the Kuramoto model can manifest stable limit cycles—periodic orbits characterized by complex patterns that the system trajectories converge to. Furthermore, in the case of linear GRAND, its flow map remains linear, effectively a composite of linear maps. In contrast, KuramotoGNN introduces a nonlinear flow map that significantly enhances its expressive capabilities.

It's worth noting that our utilization of the standard form of the Kuramoto model represents just one facet of its potential. Different formulations of the Kuramoto equations can give rise to various dynamics, offering the opportunity for exploration beyond the standard model. For example, incorporating individual coupling strengths can introduce attraction-repulsion dynamics, potentially leading to cluster synchronization phenomena.

In summary, our work primarily provides a perspective on the oversmoothing phenomenon and its connection to synchronization, with the aim of enhancing our understanding of the behavior of graph neural networks. We hope this clarification sheds light on the motivation behind our approach and its valuable contributions to the field.

\section{DATASET}
\label{appendix:data}
The statistics of the datasets are summarized in Table \ref{tab:dataset}.

\textbf{Cora} \citep{Cora}. A scientific paper citation network dataset consists of 2708 publications which are classified into one of seven classes. The citation network consists of 5429 links; each publication is represented by a vector of 0/1-valued indicating the absence/presence of the 1433 words in a corpus.

\textbf{Citeseer} \citep{Citeseer}. Similar to Cora, Citeseer is another scientific publications network consists of 3312 publications and each publication is classified into one of 6 classes. The publication is represented by a vector of 0/1 valued that also indicates the absence or presence of the corresponding word from a dictionary of 3703 unique words.

\textbf{Pubmed} \citep{Pubmed}. The Pubmed dataset consists of 19717 scientific publications related to diabetes, and all publications in the dataset are taken from the Pubmed database. Each publication is classified into one of 3 classes. The network has 44338 links and each publication is represented by TF/IDF weighted word vector from a dictionary consists of 500 unique words.

\textbf{CoauthorCS} \citep{Coauthor}. The CoauthorCS is a co-authorship graph of authors with publications related to the field of computer science. The dataset is based on the Microsoft Academic Graph from the KDD Cup 2016 challenge. In this dataset, nodes represent the authors and an edge is established if they are co-authored in a paper. Each node is classified into one of 15 classes, and each node is represented by a vector of size 6805 indicating the paper keywords for each author's papers. The network consists of 18333 nodes and 163788 edges.

\textbf{Computers} \citep{ComPhoto}. Computers dataset is a segment of the Amazon co-purchase graph. In this graph, each node is classified into one of 10 classes and each node is represented as a product. If two products are often bought together, an edge will be established. Each product is represented by a bag-of-words features vector of size 767. The data set consists of a total of 13752 products and 491722 relations between two products.

\textbf{Photo} \citep{ComPhoto}. Similar to Computers, Photo is another segment of Amazon co-purchase graph, the properties of nodes and edges are exactly the same with Computers. In this dataset, the network consists of 238163 edges and 7650 nodes, each node is classified into one of eight classes and each node is represented by a vector size of 745. 

{\setlength{\tabcolsep}{0.7cm}%
\begin{table*}[htb]
\centering
    \caption{Statitics of 6 datasets}
    
    \begin{tabular}{ccccc}
         \hline
         \hline
         Dataset & Classes & Features & \#Nodes & \#Edges\\
         \hline
          CORA & 7 & 1433 & 2485 & 5069 \\
          Citeseer & 6 & 3703 & 2120 & 3679 \\
          Pubmed & 3 & 500 & 19717 & 44324 \\
          CoauthorCS & 15 & 6805 & 18333 & 81894 \\
          Computer & 10 & 767 & 13381 & 245778 \\
          Photo & 8 & 745 & 7487 & 119043 \\
         \hline
         \hline
    \end{tabular}
      \newline
    \label{tab:dataset}
\end{table*}
}

\section{DETAIL ON GRAPH NEURAL DIFFUSION}
We recall that the Graph Neural Diffusion (GRAND) is governed by the following Cauchy problem.

\begin{gather}
    \label{diffusion}
    \frac{{\rm d} \mathbf{X}(t)}{{\rm d}t} = \div(\mathcal{G}(X(t),t) \odot \nabla \mathbf{X}(t))  \\
    \mathbf{X}(0) = \psi(\mathbf{V})
\end{gather}

Where $d$ is the size of encoded input features, $\psi:\mathbb{R}^{n\times f}\rightarrow\mathbb{R}^{n\times d}$ is an affine map that represents the encoder function to the input node features, $\mathbf{X(t)}=[\bigl(\mathbf{x}_1(t)\bigr)^\top, \dots, \bigl(\mathbf{x}_n(t)\bigr)^\top]^\top \in \mathbb{R}^{n \times d}$ is the node function matrix, $\odot$ is the point-wise multiplication, and $\div$ is the divergence operator. 

The gradient of a node-function matrix $\mathbf{X}$ is an edge-function $\nabla \mathbf{X} \in \mathbb{R}^{n \times n \times d}$ with $[\nabla \mathbf{X}]_{ij}=\mathbf{x}_j-\mathbf{x}_i \in \mathbb{R}^d$. And $\mathcal{G}=(\mathbf{X}(t),t) \in \mathbb{R}^{n \times n}$ is a matrix function which takes $\mathbf{X}(t)$ as input of the function. Furthermore, $\mathcal{G}$ always satisfies the condition that each row of $\mathcal{G} \odot \mathbf{E}$ summing to 1. Finally, $\div$ or the divergence of an edge function $\nabla \mathbf{X}$, $\div(\nabla \mathbf{X})=([\div(\nabla \mathbf{X})]_1^\top,\dots,[\div(\nabla \mathbf{X})]_n^\top) \in \mathbb{R}^{n \times d}$ is defined as:
\begin{equation}
    [\div(\nabla \mathbf{X})]_i = \sum_{i=1}^n E_{ij}[\nabla \mathbf{X}]_{ij}
\end{equation}

\section{TRAINING OBJECTIVE}
The full training optimizes the cross-entropy loss:
\begin{equation}
    \mathcal{L}(\mathbf{Y},\mathbf{T}) = H(\mathbf{Y},\mathbf{T}) = \sum_{i=1}^n \mathbf{t_i}^\top\log \mathbf{y}_i
\end{equation}
where $\mathbf{t}_i$ is the one-hot truth vector of the $i^{th}$ node and $\mathbf{y}_i=\phi(\mathbf{x}_i(T))$ is the prediction of the KuramotoGNN with $\phi : \mathbb{R}^d \rightarrow \mathbb{R}^{num\_class}$ is a linear decoder function:
\begin{align}
    \mathbf{y}_i &= \mathbf{D}\mathbf{x}_i(T)+b\\
     &= \mathbf{D}\left(\mathbf{x}_i(0)+\int_0^T \frac{{\rm d} \mathbf{x}_i(t)}{{\rm d}t}dt\right)+b\\
     &= \mathbf{D}\left(\psi(\mathbf{V_i})+\int_0^T \frac{{\rm d} \mathbf{x}_i(t)}{{\rm d}t}dt\right)+b\\
     &= \mathbf{D}\left(\mathbf{M}\mathbf{V_i}+b_{\psi}+\int_0^T \frac{{\rm d} \mathbf{x}_i(t)}{{\rm d}t}dt\right)+b
\end{align}
Moreover, $T$ is the terminated time of the ODE and $\frac{{\rm d} \mathbf{x}_i(t)}{{\rm d}t}$ is the KuramotoGNN equation which is the below equation. Note that $\mathbf{V}_i$ is the input feature vector of the $i^{th}$ node and $\mathbf{M},\mathbf{D},\frac{{\rm d} \mathbf{x}_i(t)}{{\rm d}t}$ contains learnable parameters.

\begin{equation}
    \frac{{\rm d} \mathbf{x}_i(t)}{{\rm d}t} =\omega_i + K\sum_{j \in \mathcal{N}(\mathbf{x}_i)} a_{ij}\sin(\mathbf{x}_j-\mathbf{x}_i)
\end{equation}
In here, $\omega_i=\mathbf{x}_i(0)=\mathbf{M}\mathbf{V_i}$, and $a_{ij}=softmax\left(\frac{\mathbf{W}_K\mathbf{X}(0)(\mathbf{W}_Q\mathbf{X}(0))^\top}{\sqrt{d_k}} \right)$ with $d_k$ is a constant, $\mathbf{W}_K, \mathbf{W}_Q$ are learnable parameters, and $K$ is the coupling strength constant.

\section{EXPERIMENTAL DETAILS AND MORE RESULTS}
For solving the ODEs, we use torchdiffeq library ODE Solver \citep{NODE}. For the encoder $\psi$, we employ a simple fully connected layer with dropout. Also for the decoder, after obtaining results from solving the ODEs, $X(T)$, we pass it through a simple fully connected layer to get final labels.

For all six graph node classification datasets, including CORA, CiteSeer, PubMed, coauthor graph CoauthorCS, and Amazon co-purchasing graphs Computer and Photo, we consider the largest connected component. Table \ref{tab:fine_tune_T} lists the fine-tuned $T$, and Table \ref{tab:fine_tune_K} lists the fine-tuned coupling strength $K$ for the results in the main paper. 

Although our $T$ values are smaller, please note that we use a non-linear interaction function $\sin$ instead of a linear interaction function $f(\mathbf{x})=\mathbf{x}$. That indicates our model requires more iterations for the ode solver to solve it, and each iteration is equivalent to a layer of neural network. Therefore, in terms of "real depth", our model is still deeper than GRAND-l. Table \ref{tab:num_iteration} shows the iterations in one epoch (we used the adaptive solver dopri5) of our model compared to GRAND-l. 

\begin{table}[htb]
\centering
    \caption{Comparing solver iterations for KuramotoGNN and GRAND's ODE equation.}

    \begin{tabular}{ccccccc}
         \hline
         \hline
         Model & CORA & Citeseer & Pubmed & CoauthorCS & Computer & Photo\\
         \hline
         KuramotoGNN&1900 & 1200 & 120 & 100 & 115 & 85\\
         \hline
         GRAND-l&200 & 300 & 50 & 50 & 100 & 70\\
         \hline
         \hline
    \end{tabular}
    \label{tab:num_iteration}
\end{table}

\begin{table}[htb]
\centering
    \caption{Fine-tuned $T$ for KuramotoGNN and GRAND-l.}

    \begin{tabular}{ccccccc}
         \hline
         \hline
         Model & CORA & Citeseer & Pubmed & CoauthorCS & Computer & Photo\\
         \hline
         KuramotoGNN&12&5&8&0.8&1&1.5\\
         \hline
         GRAND-l&18.2948&7.8741&12.9423&3.2490&3.5824&3.6760\\
         \hline
         \hline
    \end{tabular}
    \label{tab:fine_tune_T}
\end{table}

\begin{table*}[t]
\centering
    \caption{Mean and std of classification accuracy of KuramotoGNN and other GNNs with different number of labeled data per class (\#per class) on six benchmark graph node classification tasks. The highest accuracy is highlighted in bold for each number of labeled data per class. (Unit: \%)}
    \resizebox{\textwidth}{!}{\begin{tabular}{ccccccccc}
         \hline
          Model & \#per class & CORA & Citeseer & Pubmed & Computers & CoauthorCS & Photo \\  
         \hline 
         &1& \textbf{63.48$\pm$7.2} & \textbf{62.06$\pm$4.55} &\textbf{65.93$\pm$3.65} & 62.26$\pm$7.73 & 60.48$\pm$2.7 & 80.18$\pm$1.8\\
         &2& \textbf{71.17$\pm$5.0} & \textbf{66.85$\pm$6.72} & \textbf{72.62$\pm$3.15} & 76.24$\pm$2.72 & 75.89$\pm$0.73 & 82.67$\pm$0.8\\
         KuramotoGNN&5& \textbf{79.11$\pm$0.91}& \textbf{72.42$\pm$2.0} &76.43$\pm$1.73& 81.43$\pm$0.78 & 87.22$\pm$0.99 & \textbf{89.35$\pm$0.29}\\
         &10& \textbf{83.53$\pm$1.36} & \textbf{74.27$\pm$1.5} & 76.86$\pm$2.17& \textbf{83.84$\pm$0.54}& \textbf{90.49$\pm$0.28} & \textbf{91.35$\pm$0.1}\\
         & 20 & \textbf{85.18$\pm$1.3} & \textbf{76.01$\pm$1.4} & \textbf{80.15$\pm$0.3} & \textbf{84.6$\pm$0.59} & \textbf{92.35$\pm$0.2} & \textbf{93.99$\pm$0.17} \\
         \hline 
         &1& 54.94$\pm$16.0 & 58.95$\pm$9.59 & 65.94$\pm$4.87 & \textbf{67.65$\pm$0.37}& 60.30$\pm$1.5 & \textbf{83.12$\pm$0.78}\\
         &2& 66.92$\pm$10.04 & 64.98$\pm$8.31 & 69.31$\pm$4.87 & 76.47$\pm$1.48& 76.53$\pm$1.85 & \textbf{87.31$\pm$0.9}\\
         GRAND++&5& 77.80$\pm$4.46 & 70.03$\pm$3.63 & 71.99$\pm$1.91 & \textbf{82.64$\pm$0.56}& 84.83$\pm$0.84 & 88.33$\pm$1.21\\
         &10& 80.86$\pm$2.99 & 72.34$\pm$2.42 & 75.13$\pm$3.88 & 82.99$\pm$0.81& 86.94$\pm$0.46 & 90.65$\pm$1.19\\
         & 20 & 82.95$\pm$1.37 & 73.53$\pm$3.31 & 79.16$\pm$1.37 & 82.99$\pm$0.81 & 90.80$\pm$0.34& 93.55$\pm$0.38 \\
        \hline
         &1& 54.14$\pm$11.0 & 50.58$\pm$17.3 & 55.47$\pm$12.5 &47.96$\pm$1.3& 58.1$\pm$4.6 &76.89$\pm$2.25\\
         &2& 68.56$\pm$9.1 & 57.65$\pm$13.2 & 69.71$\pm$7.01 &75.47$\pm$1.7& 75.2$\pm$4.2 & 80.54$\pm$2.3\\
         GRAND-l&5& 77.52$\pm$3.1 & 67.48$\pm$4.2 & 70.17$\pm$4.52 & 81.23$\pm$0.6& 85.27$\pm$2.1 & 88.58$\pm$1.7\\
         with $X(0)$&10& 81.9$\pm$2.4 & 71.7$\pm$7.3 & 77.37$\pm$2.31 & 82.71$\pm$1.5& 87.6$\pm$1.8 & 90.95$\pm$0.6\\
         & 20 & 82.46$\pm$1.64 & 73.4$\pm$5.05 & 78.8$\pm$1.63 & 84.27$\pm$0.6 & 91.24$\pm$0.4& 93.6$\pm$0.4 \\
        \hline
         &1& 58.64$\pm$9.2 & 56.44$\pm$8.4 & 58.18$\pm$7.5 &48.46$\pm$10.3& \textbf{70.49$\pm$6.35} &42.02$\pm$1.9\\
         &2& 64.5$\pm$6.4 & 53.61$\pm$8.7 & 65.05$\pm$4.09 &71.29$\pm$3.4& \textbf{83.13$\pm$1.6} & 61.66$\pm$6.4\\
         GCNII&5& 76.22$\pm$0.88 & 69.2$\pm$0.9 & 70.24$\pm$0.63 & 73.60$\pm$2.1& \textbf{89.02$\pm$0.8} & 83.31$\pm$2.1\\
         &10& 75.35$\pm$1.1 & 66.29$\pm$1.2 & \textbf{76.63$\pm$1.2} & 77.83$\pm$3.9& 89.31$\pm$0.25 & 90.2$\pm$0.8\\
         & 20 & 84.02$\pm$0.5 & 70.26$\pm$0.7 & 78.95$\pm$0.9 & 80.28$\pm$2.1 & 91.11$\pm$0.2& 92.1$\pm$0.4 \\
        \hline
        &1 & 47.72$\pm$15.33 & 48.94$\pm$10.24 & 58.61$\pm$12.83 & 49.46$\pm$1.65 & 65.22$\pm$2.25 & 82.94$\pm$2.17\\
        &2 & 60.85$\pm$14.01 & 58.06$\pm$9.76 & 60.45$\pm$16.20 & \textbf{76.90$\pm$1.49} & 83.61$\pm$1.49 & 83.61$\pm$0.71\\
        GCN&5 & 73.86$\pm$7.97 & 67.24$\pm$4.19 & 68.69$\pm$7.93 & 82.47$\pm$0.97 & 86.66$\pm$0.43 & 88.86$\pm$1.56\\
        &10 & 78.82$\pm$5.38 & 72.18$\pm$3.48 & 72.59$\pm$3.19 & 82.53$\pm$0.74 & 88.60$\pm$0.50 & 90.41$\pm$0.35\\
        & 20 & 82.07$\pm$2.03 & 74.21$\pm$2.90 & 76.89$\pm$3.27 & 82.94$\pm$1.54 & 91.09$\pm$0.35 & 91.95$\pm$0.11 \\
        \hline
        &1 & 47.86$\pm$15.38 & 50.31$\pm$14.27 & 58.84$\pm$12.81 & 37.14$\pm$7.87 & 51.13$\pm$5.24 & 73.58$\pm$8.15\\
        &2 & 58.30$\pm$13.55 & 55.55$\pm$9.19 & 60.24$\pm$14.44 & 65.07$\pm$8.86 & 63.12$\pm$6.09 & 76.89$\pm$4.89\\
        GAT&5 & 71.04$\pm$5.74 & 67.37$\pm$5.08 & 68.54$\pm$5.75 & 71.43$\pm$7.34 & 71.65$\pm$4.56 & 83.01$\pm$3.64\\
        &10 & 76.31$\pm$4.87 & 71.35$\pm$4.92 & 72.44$\pm$3.50 & 76.04$\pm$0.35 & 74.71$\pm$3.35 & 87.42$\pm$2.38\\
        & 20 & 80.04$\pm$2.54 & 72.02$\pm$2.82 & 74.55$\pm$3.09 & 79.98$\pm$0.96 & 91.33$\pm$0.36 & 91.29$\pm$0.67 \\
        \hline
        &1 & 43.04$\pm$14.01 & 48.81$\pm$11.45 & 55.53$\pm$12.71 & 27.65$\pm$2.39 & 61.35$\pm$1.35 & 45.36$\pm$7.13\\
        &2 & 53.96$\pm$12.18 & 54.39$\pm$11.37 & 58.97$\pm$12.65 & 42.63$\pm$4.29 & 76.51$\pm$1.31 & 51.93$\pm$4.21\\
        GraphSAGE & 5 & 68.14$\pm$6.95 & 64.79$\pm$5.16 & 66.07$\pm$6.16 & 64.83$\pm$1.62 & 89.06$\pm$0.69 & 78.26$\pm$1.93\\
        &10 & 75.04$\pm$5.03 & 68.90$\pm$5.08 & 70.74$\pm$3.11 & 74.66$\pm$1.29 & 89.68$\pm$0.39 & 84.38$\pm$1.75\\
        & 20 & 82.07$\pm$2.03 & 71.52$\pm$4.11 & 76.49$\pm$1.75 & 73.66$\pm$2.87 & 90.31$\pm$0.41 & 88.61$\pm$1.18 \\
        \hline
    \end{tabular}}
    \label{tab:result_acc}
\end{table*}

\begin{figure}
  \centering
    \includegraphics[width=\columnwidth]{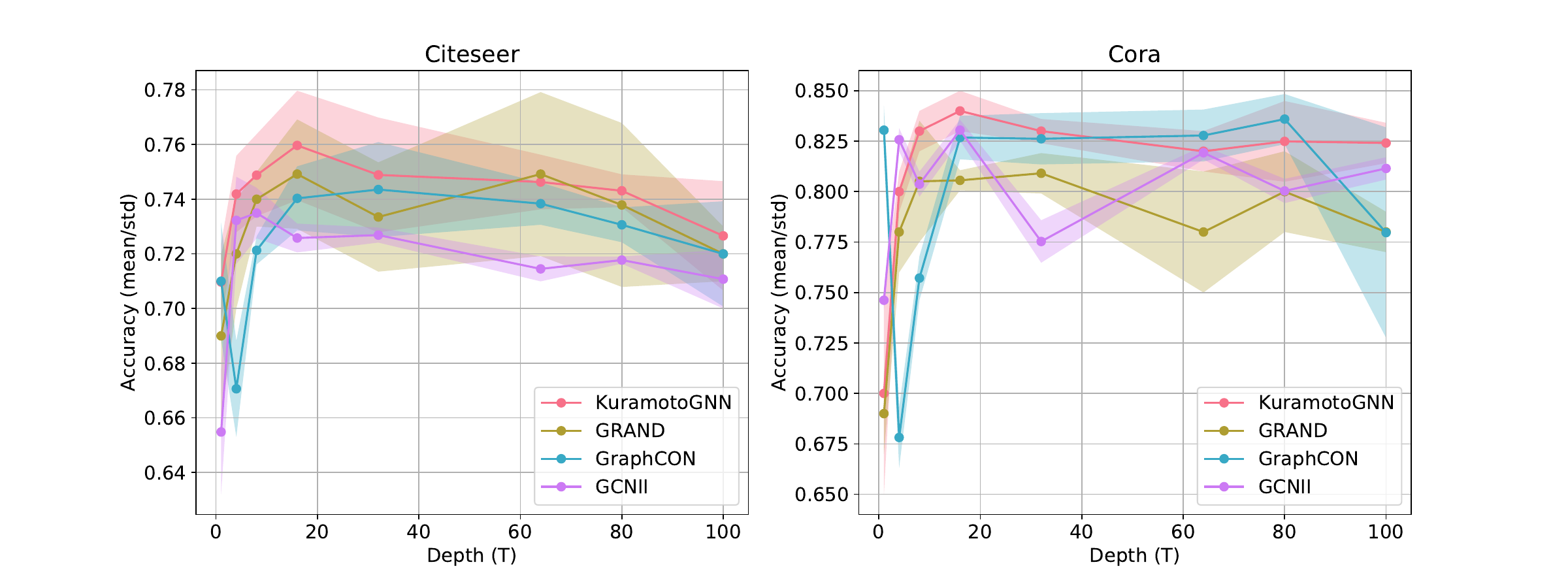}
    \centering
    \caption{Comparisions between KuramotoGNN and GNNs architectures that specifically tackle over-smoothing in different depth.}
    \label{fig:depth_2}
\end{figure}

To further test the resilience to depth, we compare KuramotoGNN with other GNNs architectures that specifically tackle over-smoothing in Table \ref{fig:depth_2} with different $T={1,4,8,16,32,64,80,100}$. Again, we used fixed-step solver Euler with step size 0.1 for fair comparison in computational process for all continuous model, except for GCNII \citep{GCNII} which is already a discretized model.

We also further explore the effects of the depth and coupling strength for KuramotoGNN by conducting further experiements based on various depths and coupling strengths. Table \ref{tab:diff_depth} shows the performances in accuracy of KuramotoGNN on three datasets: CORA, Citeseer, and Pubmed. Overall, the coupling strength is more sensitive in case of small depths, but in larger depths, the chances in performances are not significant between choices of coupling strengths.

\begin{table*}[!htb]
\centering
    \caption{Mean and std of classification accuracy of KuramotoGNN in different depths and coupling strengths on three CORA, Citeseer, and Pubmed graph node classification tasks. (Unit: \%)}
    \begin{tabular}{ccccc}
         \hline
          Depth $T$ & Coupling Strength $K$ & CORA & Citeseer & Pubmed  \\  
         \hline 
         &0.7& 75.13$\pm$1.35 & 70.24$\pm$3.41 &76.81$\pm$1.69\\
         2 &0.8& 76.27$\pm$2.89 & 71.85$\pm$2.16 & 78.07$\pm$1.77\\
         &0.9& 79.8$\pm$0.77& 73.87$\pm$1.63 &79.85$\pm$1.19\\
         &1& 78.15$\pm$0.98 & 70.89$\pm$1.81 & 77.61$\pm$2.22\\
         \hline 
         &0.7& 75.89$\pm$1.77 & 72.42$\pm$2.26 &76.88$\pm$2.58\\
         3 &0.8& 78.43$\pm$1.08 & 76.33$\pm$2.48 & 79.34$\pm$1.48\\
         &0.9& 79.67$\pm$0.77 & 72.58$\pm$2.91 &78.77$\pm$0.99\\
         &1& 79.54$\pm$2.14 & 74.8$\pm$1.19 & 79.13$\pm$0.99\\
        \hline
         &0.7& 82.03$\pm$1.79 & 72.54$\pm$1.31 & 77.98$\pm$2.95\\
         5 &0.8& 79.37$\pm$0.49 & 72.58$\pm$2.77 & 79.46$\pm$0.48\\
         &0.9& 81.98$\pm$1.96& 74.88$\pm$1.22 &78.91$\pm$2.47\\
         &1& 82.92$\pm$0.88 & 73.99$\pm$0.84 & 80.46$\pm$1.82\\
        \hline
         &0.7& 81.37$\pm$1.13 & 74.56$\pm$1.65 &80.17$\pm$0.80\\
         8 &0.8& 82.49$\pm$0.74 & 75.24$\pm$2.39 & 79.75$\pm$1.28\\
         &0.9& 82.77$\pm$1.29& 75.4$\pm$2.43 &80.07$\pm$0.57\\
         &1& 83.22$\pm$1.57 & 75.04$\pm$0.7 & 78.49$\pm$3.03\\
        \hline
         &0.7& 82.26$\pm$1.05 & 74.4$\pm$3.4 & 79.49$\pm$1.16\\
         10 &0.8& 82.49$\pm$0.98 & 75.93$\pm$1.18 & 79.27$\pm$0.52\\
         &0.9& 81.6$\pm$0.98& 74.56$\pm$0.77 &78.17$\pm$1.86\\
         &1& 83.43$\pm$1.3 & 75.12$\pm$1.02 & 79.08$\pm$1.93\\
        \hline
         &0.7& 83.53$\pm$0.72 & 74.88$\pm$1.87 & 78.84$\pm$1.69\\
         12 &0.8& 83.83$\pm$0.59 & 75.93$\pm$1.44 & 79.9$\pm$0.95\\
         &0.9& 81.75$\pm$1.61& 74.35$\pm$1.99 & 79.93$\pm$0.52\\
         &1& 85.18$\pm$1.35 & 73.63$\pm$1.72 & 79.35$\pm$0.8\\
        \hline
         &0.7& 84.06$\pm$2.46 & 73.55$\pm$0.96 & 77.49$\pm$1.41\\
         16 &0.8& 82.06$\pm$2.05 & 74.68$\pm$1.17 & 78.67$\pm$1.36\\
         &0.9& 83.35$\pm$0.48& 76.01$\pm$1.45 & 75.77$\pm$0.12\\
         &1& 82.82$\pm$1.13 & 75.16$\pm$1.19 & 75.51$\pm$2.18\\
        \hline
         &0.7& 84.37$\pm$0.68 & 75.16$\pm$0.94 &78.46$\pm$2.46\\
         18 &0.8& 82.97$\pm$0.75 & 75.28$\pm$1.21 & 76.60$\pm$2.15\\
         &0.9& 82.99$\pm$0.44& 73.83$\pm$1.95 &75.67$\pm$1.63\\
         &1& 82.79$\pm$0.38 & 75.4$\pm$1.49 & 74.2$\pm$1.71\\
        \hline

    \end{tabular}
      \vspace{0.3cm}
    \label{tab:diff_depth}
\end{table*}

\begin{figure*}[!ht]
  \centering
    \includegraphics[width=\textwidth]{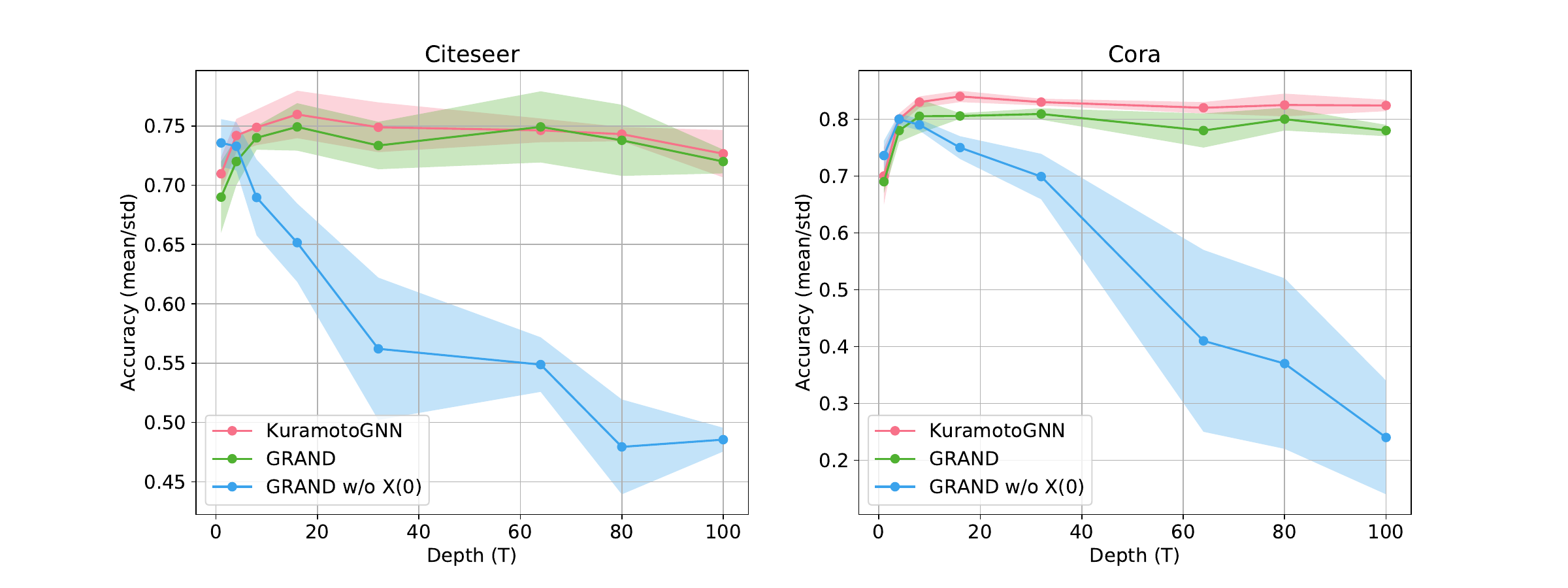}
    \centering
    \caption{Change in performance at different depth (T) on Cora and Citeseer dataset.}
  \label{fig:depth}    
\end{figure*}


\subsection{EVALUATIONS ON LIMITED TRAINING DATA}
Besides helping to avoid over-smoothing and being able to train in deep layers, KuramotoGNN also can boost the performance of different tasks with low-labeling rates. Table \ref{tab:result_acc} compares the accuracy of KuramotoGNN with other GNNs. We notice that with few labeled data, in most tasks, KuramotoGNN is significantly more accurate than the other GNNs including GRAND-l. Only for CoauthorCS and Photo datasets, the GCN outperforms both KuramotoGNN and GRAND-l on extreme limited label cases.

\subsection{EVALUATIONS ON HETEROPHILIC DATASET}
To further demonstrate the effectiveness of KuramotoGNN, we include more experiments on the node classification task using heterophilic graph datasets: Cornell, Texas and Wisconsin from the CMU WebKB\footnote{\url{http://www.cs.cmu.edu/afs/cs.cmu.edu/project/theo-11/www/wwkb/}.} project. The edges in these graphs represent the hyperlinks between webpages nodes. The labels are manually selected into five classes, student, project, course, staff, and faculty. The features on node are the bag-of-words of the web pages.The 10 generated splits of data are provided by \cite{pei2020geom}. 

{\setlength{\tabcolsep}{0.5cm}%
\begin{table}[htb]
\centering
    \caption{fine-tuned coupling strength $K$ for KuramotoGNN.}

    \begin{tabular}{cc}
         \hline
         \hline
         Dataset & Coupling strength $K$\\
         \hline
         CORA & 1\\ 
         Citeseer & 2\\
         Pubmed & 0.9\\
         CoauthorCS & 1.8\\
         Computer & 4\\
         Photo & 2\\
         \hline
         \hline
    \end{tabular}
    \label{tab:fine_tune_K}
\end{table}
}

Table \ref{tab:hetero} shows the performances of KuramotoGNN when compare with other differential equations based models: GRAND-l, GraphCON and discretized model: GCNII \citep{GCNII} . All the baselines are proceduced/re-proceduced from public code. 


We also conducted experiments on two recent challenging heterophilic datasets: \textbf{roman-empire} and \textbf{amazon-ratings}. The experimental results have been included in Table \ref{tab:hetero_2} for GRAND-l, GraphCON, and KuramotoGNN. 

Notably, the new heterophilic datasets appear to present challenges for all models, and we believe that further investigation into their dynamics and characteristics is needed. We'd like to highlight that, similar to GRAND-l, GraphCON employs a task-specific residual trick in its code. This trick's impact can vary, proving beneficial for some datasets while potentially negatively affecting others. We have thoroughly explored both versions of GraphCON, with and without this trick, to provide a comprehensive comparison. In the table, we define \textbf{GraphCON-res} as the version we used residual trick as standard public code, while GraphCON is the version we remove the trick. The same notion goes for KuramotoGNN and \textbf{KuramotoGNN-res}.

Additionally, we've observed that incorporating a similar trick into our model also yields performance improvements in certain datasets. This finding emphasizes the importance of considering dataset-specific characteristics when applying such techniques. In conclusion, the inclusion of these new heterophilic datasets has enabled us to broaden our insights into the performance of KuramotoGNN, GraphCON, and GRAND-l.

\begin{table}[htb!]
\centering
    \caption{Performance comparison between KuramotoGNN and GRAND, GraphCON for two new heterophlic datasets.}

    \begin{tabular}{ccccccc}
         \hline
         \hline
         Model & roman-empire & amazon-ratings\\
         \hline
        GRAND-l & 60.1$\pm$0.4   & 40.3$\pm$0.4  \\
         \hline
GraphCON          & 73.2$\pm$0.4   & 42.3$\pm$0.4  \\
         \hline
GraphCON-res      & 85.5$\pm$0.7   & 41.2$\pm$0.6     \\
         \hline
KuramotoGNN       & 83.0$\pm$0.5   & 41.9$\pm$0.4   \\
         \hline
KuramotoGNN-res & \textbf{86.07$\pm$0.6}  & \textbf{42.9$\pm$0.7}   \\ 

         \hline
         \hline
    \end{tabular}
    \label{tab:hetero_2}
\end{table}

\begin{table}[ht]
    \centering
    \caption{Classification accuracy on heterophilic graph node classification task.}
    \begin{tabular}{cccc}
         \hline
         \hline
         Model & Texas & Wisconsin & Cornell\\
         \hline
         KuramotoGNN& \textbf{85.4$\pm$6.2} & \textbf{87.6$\pm$3.3} &77.49$\pm$3.3 \\
         \hline
         GCNII & 81.08$\pm$4.5 & 82.31$\pm$3.1 & 79.7$\pm$6.7\\
         \hline
         GraphCON & 81.1$\pm$3.6 & 85.2$\pm$3.1 & \textbf{84.3$\pm$4.8}\\
         \hline
         GRAND-l & 78.11$\pm$7.4 & 80.39$\pm$5.4 & 62.97$\pm$6.8\\
         \hline
         \hline
    \end{tabular}
    \label{tab:hetero}
\end{table}

\begin{figure}
  \centering
    \includegraphics[width=0.6\columnwidth]{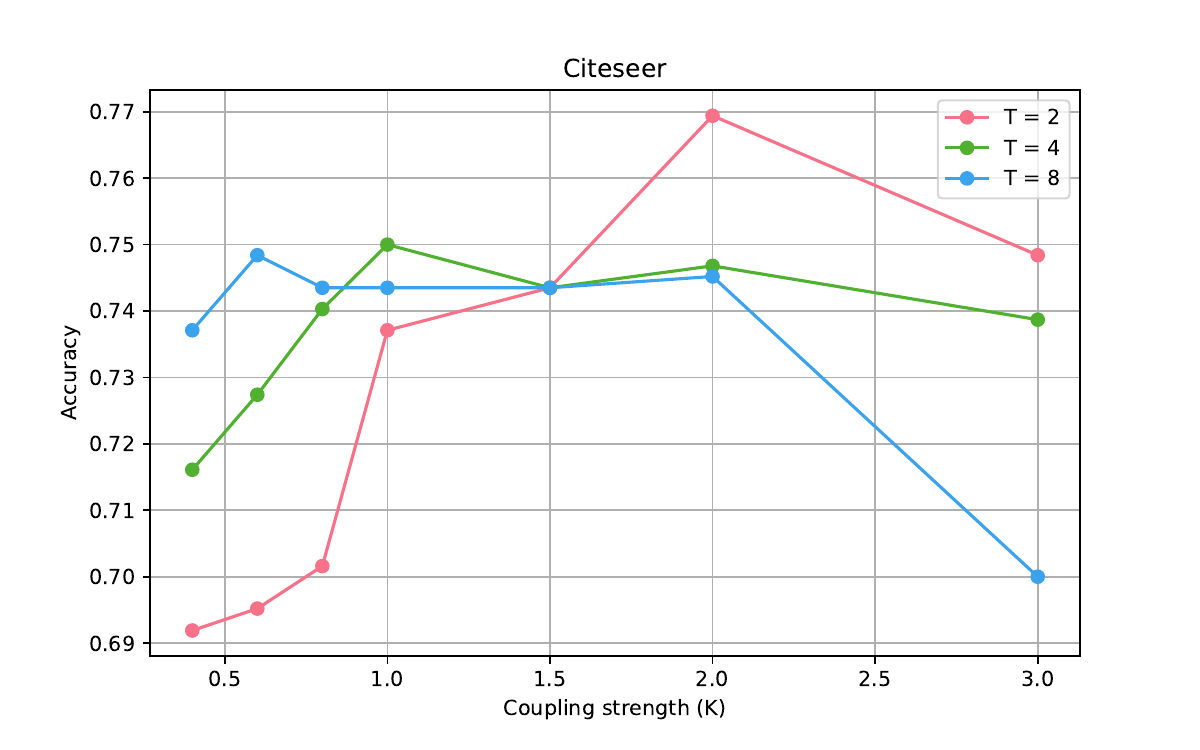}
    \centering
    \caption{Change in performance of KuramotoGNN at different coupling strength (K) on Citeseer dataset.}
    \label{fig:K}
\end{figure}

\subsection{FURTHER EVALUATING ON LARGER DATASET}

\textbf{Open graph benchmark with paper citation network (ogbn-arxiv) \citep{hu2020open}.} \quad Ogbn-arxiv consists of 169,343 nodes and 1,166,243 directed edges. Each node is an arxiv paper represented by a 128-dimensional features and each directed edge indicates the citation direction. This dataset is used
for node property prediction and has been a popular benchmark to test the advantage of deep graph neural networks over shallow graph neural networks.

We have conducted additional experiments on the OGBN-arXiv node classification task. The results in Table \ref{tab:ogbn} show that our KuramotoGNN improves over GRAND-l. We used the Euler fixed step solver with step size 0.1 for a fair comparison in the computational process.
\begin{table}[ht]
    \centering
    \caption{Classification accuracy of the GRAND-l and KuramotoGNN trained with different depth on the OGBN-arXiv graph node classification task.}
    \begin{tabular}{ccccc}
         \hline
         \hline
         Model & $T=1$ & $T=8$ & $T=32$ & $T=64$\\
         \hline
         KuramotoGNN& \textbf{66.00$\pm$0.8} & \textbf{69.87$\pm$0.2} & \textbf{69.31$\pm$0.3} & \textbf{68.32$\pm$0.2}\\
         \hline
         GRAND-l & 64.43$\pm$0.5 & 69  .02$\pm$0.4 & 67.81$\pm$0.4&66.58$\pm$1.2\\
         \hline
         \hline
    \end{tabular}
    \label{tab:ogbn}
\end{table}

\subsection{EFFECT OF COUPLING STRENGHT $K$ ON KURAMOTOGNN}
To further investigate the effect of hyper-parameter $K$ using empirical results, in the following experiments, we tried different settings of $K=\{0.4,0.6,0.8,1,1.5,2,3,\}$ on the Citeseer dataset using standard Planetoid split and on different depth $T=\{2,4,8\}$.

Figure \ref{fig:K} shows the change in performances of Kuramoto on the Citeseer dataset on different settings of $K$. It is observed that the KuramotoGNN performs well on small values of $K$, while for too small $K$, it indicates not so much change for the coupled function, and for higher $K$, the performances start decreasing. However, this phenomenon is quite well matched with the analysis of the Kuramoto model \citep{10.1007/BFb0013365, STROGATZ20001}, in which the higher the coupling strength $K$, the system tends to synchronize better. Furthermore, we also do not suggest putting $K$ too high, since it will increase the NFE (Number of Function Evaluations) of the solver to obtain an accurate solution, and thus, increasing the time of training.

\section{PROOF FOR THEOREM 4.11}
Let us recall the definition of over-smoothing and the KuramotoGNN equation from the main manuscript:
\begin{gather}
    \label{syns}
    \lim_{t\to T}\|\mathbf{x}_i(t)-\mathbf{x}_j(t)\|=0, \forall i \neq j.\\
    \label{KuramotoGNN}
    \dot{\mathbf{x}_i} = \omega_i + K\sum_{j}a_{ij}\sin(\mathbf{x}_j-\mathbf{x}_i)
\end{gather}


We prove the contrapositive. It means that if condition \eqref{syns} occurs, then $\omega_i =\omega_j, \forall i,j=1,\dots,N$.

We prove this in contradiction. We assume that equation \eqref{syns} happens and $\omega_1 \neq \omega_2$, then we will try to reach a contradiction.


Since condition \eqref{syns} happens, we substitute  it into equation \eqref{KuramotoGNN} to have the following limits:
\begin{eqnarray}\label{lims}
\lim_{t\to \infty}(\dot{\mathbf{x}_i}(t)-\omega_i)=\mathbf{0},~~i=1,2.
\end{eqnarray}
Now, for all $T \in \mathbb{Z}^+$, we apply the mean value theorem to have 
\begin{eqnarray*}
\mathbf{x}_1(T+1)-\mathbf{x}_1(T)=\dot{\mathbf{x}_1}(a_T), a_T \in (T,T+1),\\
\mathbf{x}_2(T+1)-\mathbf{x}_2(T)=\dot{\mathbf{x}_2}(b_T), b_T \in (T,T+1).
\end{eqnarray*}
Using \eqref{lims}, we get
\begin{eqnarray*}
    \mathbf{x}_1(T+1)-\mathbf{x}_1(T) \rightarrow \omega_1 \quad as \quad T \rightarrow \infty,\\
    \mathbf{x}_2(T+1)-\mathbf{x}_2(T) \to \omega_2 \quad as \quad T \to \infty.
\end{eqnarray*}
Hence, with $\mathbf{d}_{12}(t) = \mathbf{x}_1(t)-\mathbf{x}_2(t)$
\begin{eqnarray*}
\mathbf{d}_{12}(T+1)-\mathbf{d}_{12}(T) \to \omega_1-\omega_2 \neq 0~~ \quad as \quad T \to \infty,
\end{eqnarray*}
which is a clear contradiction to the fact that condition \eqref{syns} happens.

\section{PROOF FOR PROPOSITION 4.3 AND 4.4}
Our proofs are motivated by \cite{GraphCON}.
\subsection{PROPOSITION 4.3}
Let us recall the equations from the main manuscript. We consider the features of the scalar node $d=1$ for simplicity. 
\begin{gather}
    \label{layer}
    x_i^m = x_i^{m-1}+\Delta t \left( x_i^0 +\frac{1}{m}\sum \sin(x_j^{m-1}-x_i^{m-1}) \right)\\
    \mathbf{X}^0 = [x_1^0,\dots,x_n^0]^\top = \mathbf{V}*\mathbf{W}
\end{gather}
where $\mathbf{V} \in \mathbb{R}^{n \times f}$, $\mathbf{W} \in \mathbb{R}^{f \times 1}$, $\Delta t \ll 1$, $m=1,2,\dots,M$.

Moreover, we are in a setting where the learning task is for the GNN to approximate the ground truth vector $\hat{\mathbf{X}}\in \mathbb{R}^n$.
Consequently, we set up the following loss function.
\begin{equation}
    J(W)=\frac{1}{2n}\sum_{i \in \mathcal{V}}\|x_i^M-\hat{x}_i\|^2
\end{equation}
We need to compute the gradient $\partial_WJ$. Using the chain rule, we obtain the following.
\begin{gather}
    \frac{\partial J}{\partial \mathbf{W}} = \frac{\partial J}{\partial \mathbf{Z}^M}\frac{\partial {\mathbf{Z}}^M}{\partial \mathbf{Z}^1}\frac{\partial {\mathbf{Z}}^1}{\partial \mathbf{Z}^0}\frac{\partial {\mathbf{Z}}^0}{\partial \mathbf{W}}\\
    \frac{\partial {\mathbf{Z}}^M}{\partial \mathbf{Z}^1}=\prod_{i=1}^M \frac{\partial {\mathbf{Z}}^i}{\partial \mathbf{Z}^{i-1}}
    \label{total_grad}
\end{gather}

First, we find the bound of $\|\frac{\partial {\mathbf{Z}}^M}{\partial \mathbf{Z}^{1}}\|_\infty$, $\|\frac{\partial J}{\partial \mathbf{Z}^{M}}\|_\infty$, $\|\frac{\partial \mathbf{Z}^{1}}{\partial \mathbf{Z}^0}\|_\infty$,  $\|\frac{\partial \mathbf{Z}^{0}}{\partial \mathbf{W}}\|_\infty$, then we can multiply these terms together to get the final upper bound.

\begin{gather}
    \frac{\partial {\mathbf{Z}}^M}{\partial \mathbf{Z}^{M-1}} = diag(\mathbf{A})+\Delta t \mathbf{B}\\
    \mathbf{A}=\left[\begin{array}{c}
         1- \Delta t\frac{1}{n}\sum_j \cos(x_j^{N-1}-x_n^{N-1}) \\
         \vdots \\
         1- \Delta t\frac{1}{n}\sum_j \cos(x_j^{N-1}-x_n^{N-1})
    \end{array}
    \right]\\
    \mathbf{B}=\left[\begin{array}{cccc}
         0 & \frac{1}{n}\cos(x_2^{N-1}-x_1^{N-1}) & \cdots & \frac{1}{n}\cos(x_n^{N-1}-x_1^{N-1})\\
         \vdots & \vdots & \ddots & \vdots \\
          \frac{1}{n}\cos(x_1^{N-1}-x_n^{N-1}) &  \frac{1}{n}\cos(x_2^{N-1}-x_n^{N-1}) & \cdots & 0 
    \end{array}
    \right]
\end{gather}

Using the triangle inequality, we can obtain the upper bound for $\left\lVert\frac{\partial {\mathbf{Z}}^M}{\partial \mathbf{Z}^{M-1}}\right\rVert_\infty$:
\begin{gather}
    \left\lVert\frac{\partial {\mathbf{Z}}^M}{\partial \mathbf{Z}^{M-1}}\right\rVert_\infty \leq \|diag(\mathbf{A}) \|_\infty + \Delta t \|\mathbf{B}\|_\infty \leq (1+\Delta t)+\frac{\Delta t}{n}\\
    \text{Thus}, \left\lVert \frac{\partial {\mathbf{Z}}^M}{\partial \mathbf{Z}^{1}}\right\rVert_\infty \leq \left[1+(1+\frac{1}{n})\Delta t\right]^M\\
\end{gather}
With sufficiently small $\Delta t$, we have this inequality:
\begin{gather}
    \left[1+(1+\frac{1}{n})\Delta t\right]^M \leq 1+2M(1+\frac{1}{n})\Delta t
\end{gather}
leads to the following bound,
\begin{gather}
    \left\lVert \frac{\partial {\mathbf{Z}}^M}{\partial \mathbf{Z}^{1}}\right\rVert_\infty \leq 1+2M(1+\frac{1}{n})\Delta t
    \label{zn_z1}
\end{gather}
A straight-forward differentiation of $\frac{\partial J}{\partial \mathbf{Z}^M}$ yields,
\begin{gather}
    \label{4.4_j_zn}
    \frac{\partial J}{\partial \mathbf{Z}^M} = \frac{1}{n}[x_1^M -\hat{x}_1,\dots,x_n^M -\hat{x}_n]
\end{gather}
From \eqref{layer} we can easily obtain the following inequality:
\begin{gather}
    |x_i^M| \leq |x_i^{M-1}|+\Delta t(|x_i^0|+1)\\
    \text{Thus}, |x_i^M| \leq N\Delta t(|x_i^0|+1)
\end{gather}
Hence,
\begin{gather}
    \left\lVert\frac{\partial J}{\partial \mathbf{Z}^M}\right\rVert_\infty \leq \frac{1}{n}(\max|x_i^M|+\max|\overline{x}_i|) \leq \frac{1}{n}(M\Delta t\max|x_i^0|+M\Delta t+\max|\overline{x}_i|)
    \label{j_zn}
\end{gather}
Finding the bound for $\left\lVert\frac{\partial {\mathbf{Z}}^1}{\partial \mathbf{Z}^{0}}\right\rVert_\infty$ is similar to $\left\lVert\frac{\partial {\mathbf{Z}}^N}{\partial \mathbf{Z}^{1}}\right\rVert_\infty$,
\begin{gather}
    \frac{\partial {\mathbf{Z}}^1}{\partial \mathbf{Z}^0}=diag(\mathbf{C})+\Delta t \mathbf{D}\\
    \mathbf{C}=\left[\begin{array}{c}
         1- \Delta t\left(1+\frac{1}{n}\sum_j \cos(x_j^{0}-x_n^{0})\right) \\
         \vdots \\
         1- \Delta t\left(1+\frac{1}{n}\sum_j \cos(x_j^{0}-x_n^{0})\right)
    \end{array}
    \right]\\
    \mathbf{D}=\left[\begin{array}{cccc}
         0 & \frac{1}{n}\cos(x_2^{0}-x_1^{0}) & \cdots & \frac{1}{n}\cos(x_n^{0}-x_1^{0})\\
         \vdots & \vdots & \ddots & \vdots \\
          \frac{1}{n}\cos(x_1^{0}-x_n^{0}) &  \frac{1}{n}\cos(x_2^{0}-x_n^{0}) & \cdots & 0 
    \end{array}
    \right]\\
    \left\lVert\frac{\partial {\mathbf{Z}}^1}{\partial \mathbf{Z}^{0}}\right\rVert_\infty \leq \|diag(\mathbf{C})\|_\infty + \Delta t \|\mathbf{D}\|_\infty \leq 1+\frac{\Delta t}{n}
    \label{z1_z0}
\end{gather}
Then we can find a bound for \eqref{total_grad} by multiplying \eqref{j_zn}, \eqref{zn_z1}, \eqref{z1_z0} together with $\frac{\partial {\mathbf{Z}}^0}{\partial \mathbf{W}}=\mathbf{V}$,
\begin{gather}
    \left\lVert\frac{\partial J}{\partial \mathbf{W}}\right\rVert_\infty \leq \frac{1}{n}\left[\alpha(\max|x_i^0|+1)+\max|\overline{x}_i|\right](\beta+\alpha)\beta\|\mathbf{V}\|_\infty\\
    \alpha = M\Delta t, \quad \quad \beta = 1+\frac{\Delta t}{n}
\end{gather}

\subsection{PROPOSITION 4.4}
Motivated by \cite{GraphCON}, we will need the following order notation:
\begin{gather}
    \beta=\mathcal{O}(\alpha), \text{for $\alpha,\beta \in \mathbb{R}_+$ if there exists constants $\overline{C},\underline{C}$ that $\underline{C}\alpha\leq\beta\leq\overline{C}\alpha$}\\
    \mathbf{M}=\mathcal{O}(\alpha), \text{for $\mathbf{M} \in \mathbb{R}^{d_1 \times d_2},\alpha \in \mathbb{R}_+$ if there exists constants $\overline{C}$ that $\|\mathbf{M}\|\leq \overline{C}\alpha$}
\end{gather}
We can rewrite $\frac{\partial {\mathbf{Z}}^M}{\partial \mathbf{Z}^{M-1}}$ as the following
\begin{gather}
    \frac{\partial {\mathbf{Z}}^M}{\partial \mathbf{Z}^{M-1}}=\mathbf{I}+\Delta t\mathbf{E}_{M-1}\\
    \mathbf{E}_{M-1}=\left[\begin{array}{cccc}
         -\frac{1}{n} \sum_j \cos(x_j^{M-1}-x_n^{M-1}) & \frac{1}{n}\cos(x_2^{M-1}-x_1^{M-1}) & \cdots & \frac{1}{n}\cos(x_n^{M-1}-x_1^{M-1})\\
         \vdots & \vdots & \ddots & \vdots \\
          \frac{1}{n}\cos(x_1^{M-1}-x_n^{M-1}) &  \frac{1}{n}\cos(x_2^{M-1}-x_n^{M-1}) & \cdots & -\frac{1}{n}\sum_j \cos(x_j^{M-1}-x_n^{M-1}) 
    \end{array}
    \right]
\end{gather}
And then, we can calculate $\frac{\partial {\mathbf{Z}}^M}{\partial \mathbf{Z}^{1}}$
\begin{gather}
\label{4.4_zn_z1}
    \frac{\partial {\mathbf{Z}}^M}{\partial \mathbf{Z}^{1}}=\mathbf{I}+\Delta t\sum_{i=1}^M\mathbf{E}_{i-1}+\mathcal{O}(\Delta t^2)
\end{gather}
With the same manner, we can rewrite $\frac{\partial {\mathbf{Z}}^1}{\partial \mathbf{Z}^{0}}$ as
\begin{gather}
    \label{4.4_z1_z0}
    \frac{\partial {\mathbf{Z}}^1}{\partial \mathbf{Z}^{0}} = \mathbf{I}+\Delta t\mathbf{E}'\\
    \mathbf{E}'=\left[\begin{array}{cccc}
         1-\frac{1}{n} \sum_j \cos(x_j^0-x_n^0) & \frac{1}{n}\cos(x_2^0-x_1^0) & \cdots & \frac{1}{n}\cos(x_n^0-x_1^0)\\
         \vdots & \vdots & \ddots & \vdots \\
          \frac{1}{n}\cos(x_1^0-x_n^{0}) &  \frac{1}{n}\cos(x_2^{0}-x_n^{0}) & \cdots & 1-\frac{1}{n}\sum_j \cos(x_j^0-x_n^0) 
    \end{array}
    \right]
\end{gather}
Then we can obtain proposition 4.4 by multiplying \eqref{4.4_j_zn}, \eqref{4.4_zn_z1}, \eqref{4.4_z1_z0} together with $\frac{\partial {\mathbf{Z}}^0}{\partial \mathbf{W}}=\mathbf{V}$.

\section{ON THE GENERALIZATION PERFORMANCE OF KURAMOTOGNN}
Given a space $Z$ and a fixed distribution $D$ on $Z$. Let $G$ be a class of hypothesis functions: $h:Z\rightarrow \mathbb{R}^{num\_classes}$. Given a loss function, say, $l(h; z)$, whose first and second arguments are a hypothesis and input, respectively, we define $L(h)$ for the \textit{predictive loss}:
\begin{eqnarray*}
    L_D[h] = \mathbf{E}_{z\sim D}[l(h,z)]
\end{eqnarray*} 
which is the expectation of a loss function with a hypothesis $h$ over a distribution $D$ of datasets. Similarly, given a set of examples $S=(z_1,\dots,z_m)$ drawn i.i.d from $D$, writes $L_S(g)$ for the \textit{empirical loss}:
\begin{eqnarray*}
    L_S[h]=\frac{1}{m}\sum_{i=1}^m[l(h,z_i)]
\end{eqnarray*}

In statistical learning theory, our focus lies in determining the bound between estimated error (empirical loss) and true error (predictive loss) across all functions in $H$. Smaller bound is better, since it means that the true error of a classifier is not much higher than its estimated error, and so selecting a classifier that has a low estimated error will ensure that the true error is also low.

In order to finding such bound, we will need a complexity measure for the class of hypothesis functions $H$. To this end, let $\bm{\sigma}=(\sigma_1,\dots,\sigma_m)$ be a list of independent random variables, where, $P(\sigma_i=+1)=P(\sigma_i=-1)=1/2$. Then \textit{the empirical Rademacher complexity} of $l$ and $H$ with respect to $S$ is defined to be
\begin{eqnarray}
    R(l \circ H \circ S) = \mathbf{E}_{\bm \sigma}\left[ \sup_{h \in H} \frac{1}{m}\sum_{i=1}^m \sigma_i l(h,z_i) \right]    
\end{eqnarray}
Then for any integer $m \geq 1$, \textit{the Radamacher complexity} of $H$ with respect to samples size $m$ drawn according to $D$ is
\begin{eqnarray}
    R_{D,m}=\mathbf{E}_{S \sim D^m}\left[ R(l \circ H \circ S) \right]
\end{eqnarray}

\begin{remark}
    Intuitively, the empirical Rademacher complexity measures how well the class of functions $H$ correlates with randomly generated labels on the set $S$. The richer the class of functions $H$ the better the chance of finding $h \in H$ that correlates with a given $\sigma$, and hence the larger empirical Rademacher complexity.
\end{remark}

Suppose that we are given a dataset $\{(z_i,\hat{y}_i)\}_{i=1}^m$ where $m$ is the number of observable nodes in the graph, $z_i$ and $\hat{y}_i$ are the node feature vector, and label of $i^{th}$ node, respectively. $H$ is the hypothesis set of KuramotoGNN. The following Proposition indicates the generalization performance of KuramotoGNN with sufficient training sample size.
\begin{proposition}
     Given $H$ is the hypothesis set of KuramotoGNN. If a sufficiently large sample is drawn from distribution $D$, then with high probability $L_D[h]$ and $L_S[h]$ are not too far apart for all functions $h \in H$:
     \begin{eqnarray}
         \lim_{m\rightarrow +\infty} \left(L_D[h]-L_S[h]\right) = 0
     \end{eqnarray}
\end{proposition}

\begin{proof}
     The following bound holds with at least probability $1-\delta$, which is well known as the \textbf{Rademacher-based uniform convergence}.
    \begin{eqnarray}
        \label{radamacher}
        L_D[h]-L_S[h] \leq 2\mathbf{E}_{S \sim D^m}\left[R(l \circ H \circ S)\right]+c\sqrt{\frac{2\log(2/\delta)}{m}}
    \end{eqnarray}
    where $c > 0$ is a constant, $m$ is the sample size, $h \in H$. Now, we estimate the first term of the RHS. Following the training objective in Section 3 of the \textbf{SM}, we used a linear discriminator for classification. Hence, in the case of binary classification, $\mathbf{D} \in \mathbb{R}^{1 \times d}$. Hereafter, we use a notation
    \begin{equation*}
        H \circ S \equiv \{\mathbf{D}x(T;z_1)+b,\dots,\mathbf{D}x(T;z_m)+b\} \subset \mathbb{R}
    \end{equation*} with $T$ being the terminating time of the ODE. It is known that 
    \begin{eqnarray*}
        R(l \circ H \circ S) \leq \rho R(H \circ S)
    \end{eqnarray*}
    where $\rho$ is the Lipschitz coefficient of $l$. Thus, it is enough now to estimate $R(H \circ S)$. Under assumption $\|\mathbf{D}\| < +\infty$, this can be done as follows.
    \begin{align*}
        m R(H \circ S) &= \mathbf{E}_\sigma\left[\sup_{\hat{y}} \sum_{i=1}^m \sigma_i \hat{y}\right]\\
         &= \mathbf{E}_\sigma\left[\sup \sum_{i=1}^m \sigma_i (\mathbf{D}x(T;z_i)+b)\right]\\
         &\leq \|\mathbf{D}\|_\infty \mathbf{E}_\sigma\left[\sup_{\mathbf{D}}\sum_{i=1}^m|\sigma_i x(T;z_i)|\right] + \mathbf{E}_\sigma\left[\sup_b \sum_{i=1}^m\sigma_i b\right]\\
         &\leq \|\mathbf{D}\|_\infty \left(\mathbf{E}_\sigma\left[\left\lVert\sum_{i=1}^m \sigma_i x(T;z_i) \right\rVert^2\right]\right)^{\frac{1}{2}} + \sup_b b \left(\mathbf{E}_\sigma\left[ \sum_{i=1}^m\sigma_i\right]\right) \quad \quad (\text{Jensen's inequality})\\
         & \leq \|\mathbf{D}\|_\infty \left(\mathbf{E}_\sigma\left[\sum_{i=1}^m\left\lVert x(T;z_i)\right\rVert^2\right]\right)^{\frac{1}{2}}, \quad \quad (\mathbf{E}_\sigma[\sigma_i]=0)\\
         & \leq \|\mathbf{D}\|_\infty \left(m\left\lVert x_M(T;z_i)\right\rVert^2\right)^{\frac{1}{2}} \quad \quad x_M=\text{argmax }x_i
    \end{align*}
    Thus, the problem reduces to the estimate of $\|x_i(T;z_i)\|^2$. Recall that the solution to equation (10) in the main text is:
    \begin{equation*}
        x_i(T;z_i)=(1+T)x_i(0)+K\int_0^T \sum_{j=1}^n a_{ij}\sin(x_j(t)-x_i(t))\,dt
    \end{equation*}
    Together with $|a_{ij}| \leq 1$, we have
    \begin{equation*}
        \|x_i(T;z_i)\| \leq (1+T)x_i(0) + nTK
    \end{equation*}
    Combining these, we find that the RHS of \eqref{radamacher} tends to 0 as $m \rightarrow +\infty$, and therefore, with sufficient sample size, the KuramotoGNN training process is consistent with predictive loss.
\end{proof}

\section{ON THE EFFICIENCY ANALYSIS}
In this section, we embark on a thorough examination of the time and space complexity associated with the KuramotoGNN. This analysis aims to provide a comprehensive understanding of the computational efficiency and resource requirements of our proposed framework.

First, the following definition is the local order parameter which is distinct from the order parameter presented in equation (8) from the main text, which pertains to the entire system.
\begin{equation}
    r_k e^{i\phi_k}=\sum_{j}^{\mathcal{N}(k)}e^{i\theta_j}*a_{kj}
\end{equation}
where $k$ represents node $k$-th in the graph, $\mathcal{N}(k)$ is set of neighbors of $k$ and $a_{kj}$ is the row of attention matrix which sum to 1, $\phi_k=\sum_{j}^{\mathcal{N}(k)}\theta_j*a_{kj}$ is the weighted average phase of node $k$. Then we can derive:
\begin{align}
    r_k e^{i(\phi_k-\theta_k)}&=\sum_{j}^{\mathcal{N}(k)}e^{i(\theta_j-\theta_k)}*a_{kj}\\
    r_k \sin(\phi_k-\theta_k)&=\sum_{j}^{\mathcal{N}(k)}\sin(\theta_j-\theta_k)*a_{kj} \label{local_order}
\end{align}
We obtain the last line by taking the imaginary parts only. Now we can paraphrase our KuramotoGNN:
\begin{equation}
    \dot{\bm{x}_i} = \omega_i + Kr_i\sin(\bm{\phi}_i-\bm{x}_i)
\end{equation}
Now, to calculate $\mathbf{x}_i(t+\Delta t)$, we have to first calculate $\mathbf{R}=[r_1,\dots,r_n]$ and $\mathbf{\Phi}=[\phi_1,\dots,\phi_n]$. Taking the norm from \eqref{local_order} we would obtain:
\begin{align}
    |r_k| &= |\sum_{j}^{\mathcal{N}(k)}e^{i\theta_j}*a_{kj}|\\
    |r_k| &= \sqrt{(\sum_{j}^{\mathcal{N}(k)}\cos(\theta_j)*a_{kj})^2+(\sum_{j}^{\mathcal{N}(k)}\sin(\theta_j)*a_{kj})^2}
\end{align}
In terms of matrices and apply to KuramotoGNN:
\begin{align}
    |\mathbf{R}|&=\sqrt{\big(\mathbf{A}\cos(\mathbf{X})\big)^2 + \big(\mathbf{A}\sin(\mathbf{X})\big)^2}\\
    \mathbf{\Phi} &= \mathbf{A}\mathbf{X}\\
    \mathbf{X} &= \mathbf{\Omega} + K\mathbf{R}\sin(\mathbf{\Phi}-\mathbf{X})
\end{align}
where $\sin,\cos$ are element-wise. So compared to GRAND-l which has only one matrix multiplication $\mathbf{A}\mathbf{X}$ to calculate the discrete forward process, we have 3 matrix multiplications. The overall complexity analysis suggests that KuramotoGNN does not significantly deviate from GRAND-l in terms of computational complexities.

\bibliography{references}